%% file: main.tex
\documentclass{article}

\usepackage{arxiv}

\usepackage[utf8]{inputenc} 
\usepackage[T1]{fontenc}    
\usepackage{hyperref}       
\usepackage{url}            
\usepackage{booktabs}       
\usepackage{amsfonts}       
\usepackage{nicefrac}       
\usepackage{microtype}      
\usepackage{cleveref}       
\usepackage{lipsum}         
\usepackage{graphicx}
\usepackage{natbib}
\usepackage{doi}
\usepackage{multirow}
\usepackage{amssymb}
\usepackage{pifont}
\usepackage{caption}
\usepackage{subcaption}
\usepackage[table]{xcolor}

\newcommand{\BenchName}{IDEA-Bench}

\title{\BenchName: How Far are Generative Models from Professional Designing?}

\newif\ifuniqueAffiliation
\uniqueAffiliationfalse

\ifuniqueAffiliation 
\author{ \href{https://orcid.org/0000-0000-0000-0000}{\includegraphics[scale=0.06]{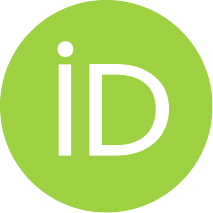}\hspace{1mm}David S.~Hippocampus}\thanks{Use footnote for providing further
		information about author (webpage, alternative
		address)---\emph{not} for acknowledging funding agencies.} \\
	Department of Computer Science\\
	Cranberry-Lemon University\\
	Pittsburgh, PA 15213 \\
	\texttt{hippo@cs.cranberry-lemon.edu} \\
	\And
	\href{https://orcid.org/0000-0000-0000-0000}{\includegraphics[scale=0.06]{orcid.pdf}\hspace{1mm}Elias D.~Striatum} \\
	Department of Electrical Engineering\\
	Mount-Sheikh University\\
	Santa Narimana, Levand \\
	\texttt{stariate@ee.mount-sheikh.edu} \\
}
\else

\author{
Chen Liang$^{2}$ \thanks{This work was done during interships at Tongyi Lab. }
\And
Lianghua Huang$^{1}$ 
\And
Jingwu Fang$^{3}$ 
\And
Huanzhang Dou$^{4}$ \footnotemark[1]
\And
Wei Wang$^{1}$
\AND
Zhi-Fan Wu$^{1}$
\And
Yupeng Shi$^{1}$
\And
Junge Zhang$^{2}$
\And
Xin Zhao$^{3}$
\And
Yu Liu$^{1}$
\AND
\\
{\large $^1$ Tongyi Lab }, {\large $^2$ CASIA }, {\large $^3$ USTB }, {\large $^4$ ZJU }
}

\renewcommand{\shorttitle}{\BenchName: How Far are Generative Models from Professional Designing?}

\hypersetup{
pdftitle={\BenchName: How Far are Generative Models from Professional Designing?},
pdfsubject={cs.CV, cs.MM, cs.AI},
pdfauthor={Chen Liang, Lianghua Huang, Jingwu Fang, Huanzhang Dou, Wei Wang, Zhi-Fan Wu, Yupeng Shi, Junge Zhang, Xin Zhao, Yu Liu},
pdfkeywords={Generative model benchmarking, Multimodal Evaluation, Task Generalization},
}

\begin{document}
\maketitle

\renewcommand{\thefootnote}{}
\footnotetext[1]{Emails: liangchen2022@ia.ac.cn, xuangen.hlh@alibaba-inc.com, u202143361@xs.ustb.edu.cn, hzdou@zju.edu.cn, \{ww413411, wuzhifan.wzf\}@alibaba-inc.com, shiyupeng.syp@taobao.com, jgzhang@nlpr.ia.ac.cn, xinzhao@ustb.edu.cn, ly103369@alibaba-inc.com.}
\renewcommand{\thefootnote}{\arabic{footnote}}

\vspace{1em}
\input{sec/0_abstract}



\input{sec/1_intro}

\input{sec/2_rela}
\input{sec/3_bench}
\input{sec/4_eval}
\input{sec/5_conclusion}

\bibliographystyle{unsrtnat}
\bibliography{references}  

\input{sec/X_suppl}






\end{document}

%% file: sec/0_abstract.tex
\vskip 0.2in
\begin{abstract}

Real-world design tasks—such as picture book creation, film storyboard development using character sets, photo retouching, visual effects, and font transfer—are highly diverse and complex, requiring deep interpretation and extraction of various elements from instructions, descriptions, and reference images. The resulting images often implicitly capture key features from references or user inputs, making it challenging to develop models that can effectively address such varied tasks. While existing visual generative models can produce high-quality images based on prompts, they face significant limitations in professional design scenarios that involve varied forms and multiple inputs and outputs, even when enhanced with adapters like ControlNets and LoRAs. To address this, we introduce \BenchName, a comprehensive benchmark encompassing 100 real-world design tasks, including rendering, visual effects, storyboarding, picture books, fonts, style-based, and identity-preserving generation, with 275 test cases to thoroughly evaluate a model’s general-purpose generation capabilities. Notably, even the best-performing model only achieves 22.48 on \BenchName, while the best general-purpose model only achieves 6.81. We provide a detailed analysis of these results, highlighting the inherent challenges and providing actionable directions for improvement. Additionally, we provide a subset of 18 representative tasks equipped with multimodal large language model (MLLM)-based auto-evaluation techniques to facilitate rapid model development and comparison. We releases the benchmark data, evaluation toolkits, and an online leaderboard at \url{https://github.com/ali-vilab/IDEA-Bench}, aiming to drive the advancement of generative models toward more versatile and applicable intelligent design systems.

\end{abstract}

%% file: sec/1_intro.tex
\section{Introduction}
\label{sec:intro}

\begin{figure}[!t]
\begin{center}
\includegraphics[width=1\linewidth]{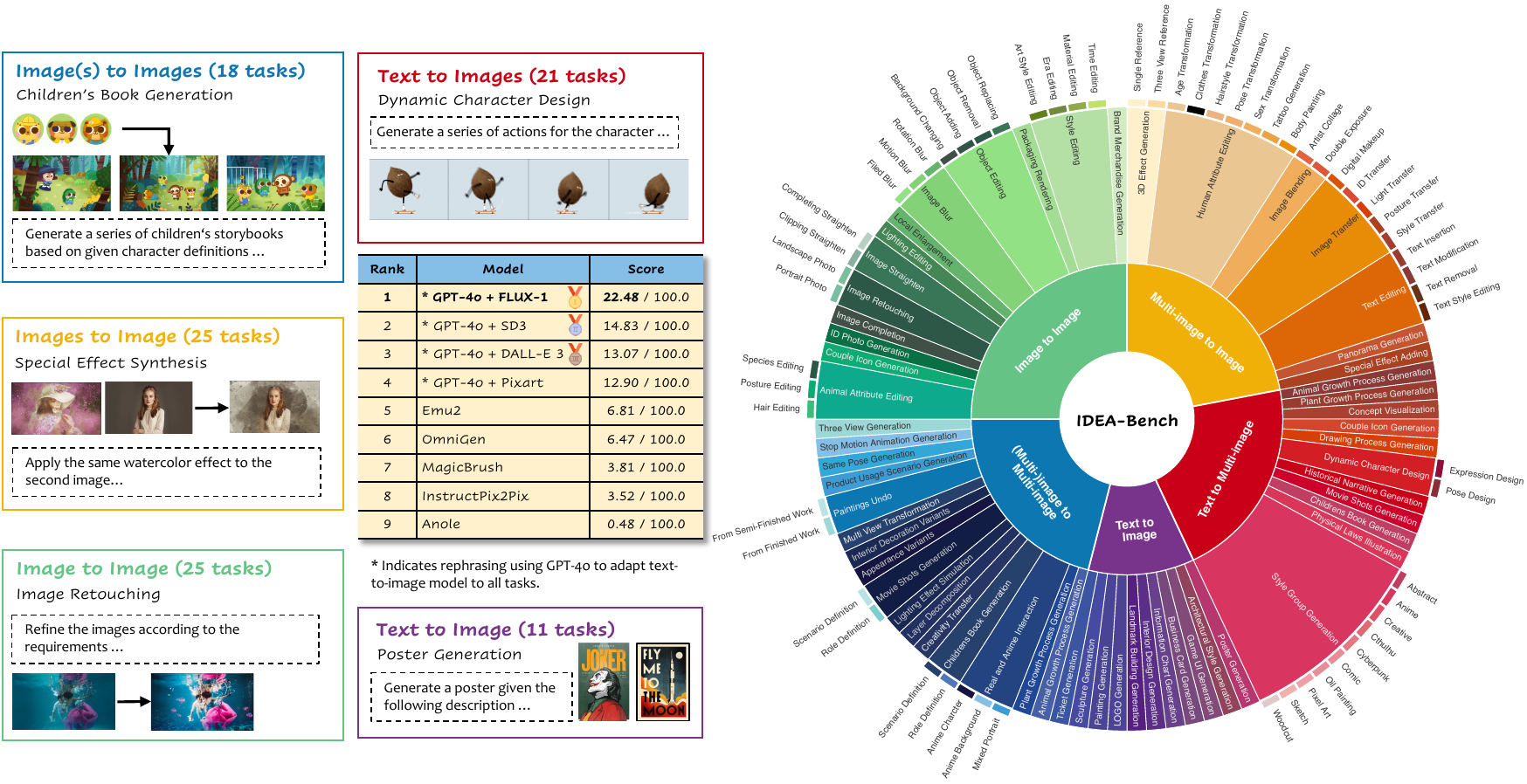}
\caption{\textbf{Overview of \BenchName.} \BenchName~comprises 5 categories, encompassing a total of 100 professional-level subtasks, 275 cases, and 1,650 hierarchical evaluation questions. Each category provides subtask examples, quantitative statistics, and showcases a leaderboard of mainstream models.} 
\label{fig:overview}
\end{center}
\end{figure}

\begin{figure}[!t]
\begin{center}
\includegraphics[width=1\linewidth]{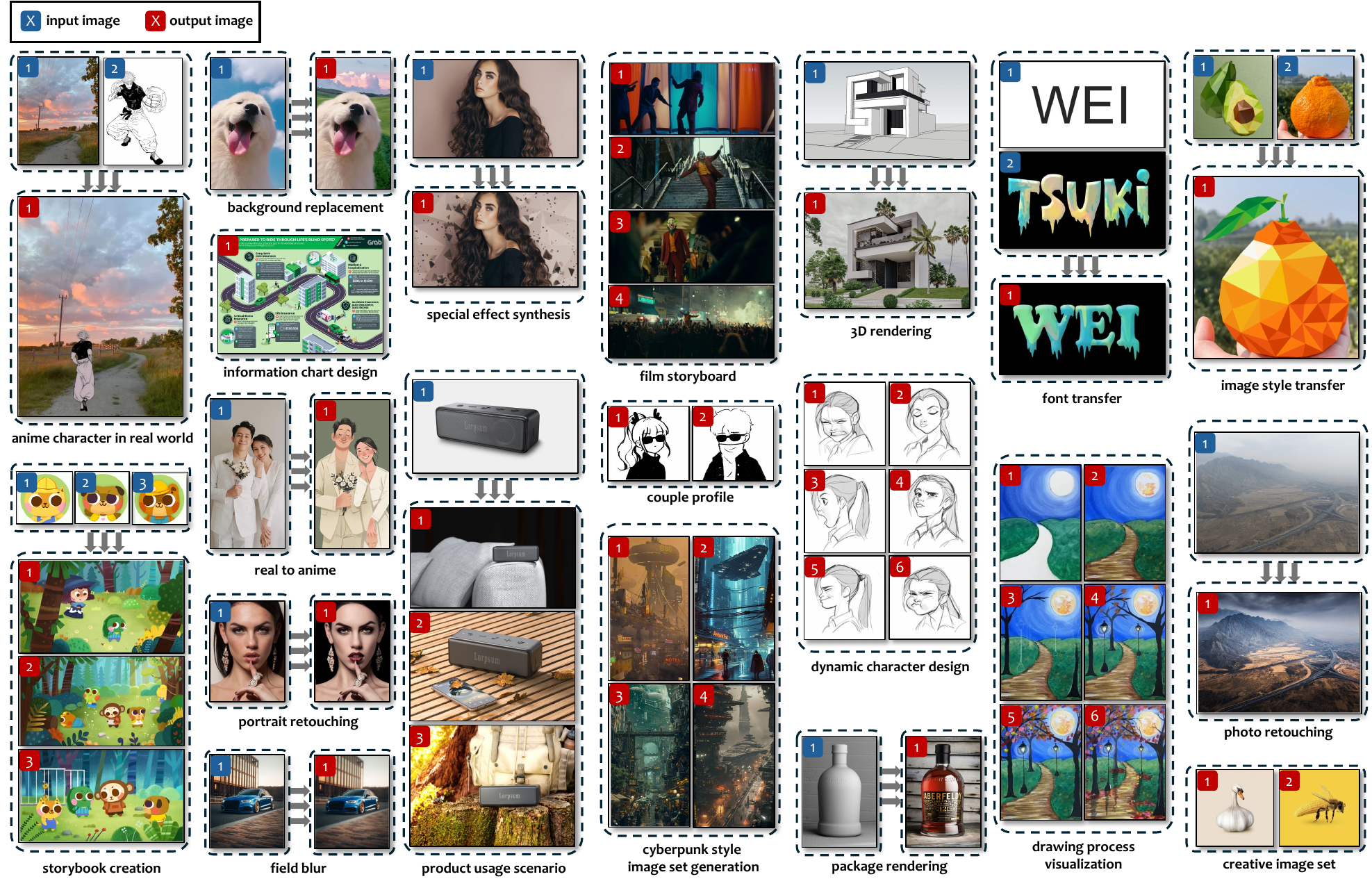}
\caption{\textbf{Demonstration of task definitions in \BenchName.} \BenchName~encompasses a diverse range of professional image design tasks. Input formats include plain text, single-image input, and multi-image input, while output formats span both single-image and multi-image generation.} 
\label{fig:vis-tasks}
\end{center}
\end{figure}

Recent advancements in Text-to-Image (T2I) models \citep{ramesh2021zero,ramesh2022hierarchical,esser2021taming,rombach2022high,saharia2022photorealistic,betker2023improving,podell2023sdxl,esser2024scaling,baldridge2024imagen,chen2023pixart,blackforestlabs_flux_2024} have significantly enhanced the ability to generate high-quality images from textual descriptions. Building on these successes, models such as ControlNet \citep{zhang2023adding} and T2I-Adapter \citep{mou2024t2i} integrate additional networks into text-to-image diffusion frameworks to incorporate visual conditioning. Similarly, InstructPix2Pix \citep{brooks2023instructpix2pix} and Emu-Edit \citep{sheynin2024emu} are specifically trained on datasets tailored for complex image editing tasks. Despite the widespread popularity of models like DALL-E 3 \citep{ramesh2022hierarchical} and FLUX-1 \citep{blackforestlabs_flux_2024}, which attract millions of visits daily, professional users often rely on established workflows and specialized software for image creation. This reliance highlights a substantial gap between the capabilities of current image generation models and the demanding requirements of professional-grade image design. When most existing image generation models focus on academic task research, the rapid evolution of large language models (LLMs) \citep{radford2019language, raffel2020exploring, brown2020language, ouyang2022training, zhang2022opt, touvron2023llama, touvron2023llama2, dubey2024llama} and multimodal language models (MLLMs) \citep{li2024llavanext-strong, wang2024qwen2, gpt4o, team2023gemini} indicate that future image generation models will increasingly aim to achieve both professional-grade quality and task unification, enabling them to handle a diverse range of complex applications seamlessly.

Recent advances in image generation have demonstrated robust task unification and generalization through large-scale pre-training and arbitrary task learning \citep{ge2023making, sun2023emu, sun2024generative, wang2024emu3, chern2024anole, huang2024group}. These models handle a wide range of academic tasks with a unified input-output format and can generalize to unseen tasks, showcasing in-context learning abilities. However, existing benchmarks \citep{cho2023dall, ghosh2024geneval, sheynin2024emu, hu2023tifa, ruiz2023dreambooth, ku2023imagenhub} are often narrow in scope, focusing primarily on isolated academic tasks and lacking comprehensive criteria to evaluate the multifaceted demands of professional image design. Consequently, there is a critical need for more robust and versatile evaluation frameworks to effectively assess the diverse and sophisticated capabilities of contemporary generative models.


To bridge the gap between generative models and professional-grade design, we introduce \BenchName~(\textbf{I}ntelligent \textbf{D}esign \textbf{E}valuation and \textbf{A}ssessment Benchmark). We conduct a thorough review of real-world design and art tasks from diverse platforms, distilling 100 representative professional image generation tasks and 275 cases that comprehensively span the essential capabilities and effects for creating all forms of artwork. These tasks are systematically categorized into five distinct types based on required capabilities: \textbf{text-to-image}, \textbf{image-to-image}, \textbf{images-to-image}, \textbf{text-to-images}, and \textbf{image(s)-to-images}, as illustrated in \cref{fig:overview}. \Cref{fig:vis-tasks} provides an intuitive display of the definition of some tasks. Currently, only a few models \citep{yang2024seed, wu24next, wang2024emu3, chern2024anole} can handle the most intricate category, which demands MLLM-level understanding of both image and text and the ability to produce variable-length image sequences. To evaluate generated images, we leverage MLLM to transform the evaluation into an image understanding task. This approach surpasses traditional metrics like FID \citep{chong2020effectively} and CLIPScore \citep{hessel2021clipscore}, which fail to capture nuances in aesthetic quality, contextual relevance, and multimodal integration. \BenchName~includes six detailed evaluation questions per case, totaling 1,650 binary scoring items, and a representative subset of 18 tasks for more nuanced and reliable assessments, ensuring precise and objective evaluations aligned with professional design standards.

Our contributions can be summarized as follows:

\begin{itemize}

\item We introduce \BenchName~to bridge the gap between current generative model capabilities and the stringent demands of professional-grade image design, which consists of 100 carefully selected tasks. 

\item We develop a task categorization of five distinct categories based on complexity and modality requirements, with a detailed evaluation framework comprising 1,650 binary scoring items, enabling precise and objective assessment.

\item We leverage MLLM-evaluation on a subset of 18 representative tasks. \BenchName~demonstrates that MLLMs have the capability to perform objective image assessment. 

\end{itemize}

%% file: sec/2_rela.tex
\section{Related Work}
\label{sec:related}

\subsection{Image Synthesis Models}

Recently, diffusion models such as Stable Diffusion \citep{podell2023sdxl, esser2024scalingrectifiedflowtransformers}, DALL-E 3 \citep{ramesh2022hierarchical}, and FLUX-1 \citep{flux1} have gained significant popularity for generating photorealistic images and offering improved training stability. However, state-of-the-art models still struggle with intricate prompts involving multiple visual concepts and extensive textual information, similar to image editing models \citep{brooks2023instructpix2pix, zhang2024magicbrush, sheynin2024emu}, which find it challenging to interpret detailed commands requiring nuanced modifications for professional-grade tasks. To enhance generative models’ capabilities, \BenchName~assesses existing image editing models against the rigorous demands of professional image design, aiming to emulate the creative and analytical processes of human designers. Additionally, image customization \citep{gal2022image, hu2021lora, ruiz2022dreambooth} is crucial for professionals needing varied personalized images. While techniques like Textual Inversion \citep{gal2022image}, LoRA \citep{hu2021lora}, and DreamBooth \citep{ruiz2022dreambooth} improve text-to-image models through fine-tuning, they lack adaptability to unseen subjects. \BenchName~aims at comprehensively evaluating models’ customization capabilities across multi-image generation tasks, assessing consistency in content, style, identity, and conceptual aspects.


\subsection{Universal Generative Model}

Universal generative models have become pivotal in advancing AI’s ability to perform a diverse array of tasks across both language and vision domains. LLMs \citep{radford2019language, raffel2020exploring, brown2020language, ouyang2022training, zhang2022opt, touvron2023llama, touvron2023llama2, dubey2024llama} have demonstrated remarkable versatility, excelling in tasks such as question answering and summarization through extensive pretraining on varied datasets. 
Building on this success, recent academic efforts \citep{team2024chameleon, zhou2024transfusionpredicttokendiffuse, sun2024generative, wang2024emu3, xiao2024omnigen} aim to adopt a more cohesive paradigm that addresses a wide range of visual generation tasks, including text-to-image and autoregressive image synthesis. 
Despite these advancements, there remains a significant gap in establishing a unified benchmark that comprehensively evaluates the multifaceted capabilities of these generative models. \BenchName~establishes a unified evaluation standard to assesses the strengths and limitations of existing generative foundation models, while guiding the development towards handling realistic and complex real-world image generation tasks at the same time.

\subsection{Benchmarks for Generative Models}

\begin{table}[!t]
\centering
\caption{\textbf{Comparison with other image generation benchmarks.} \BenchName~offers a broader range of task categories, multiple levels of model capabilities for evaluation, and longer and more complex prompts.}
\resizebox{0.8\textwidth}{!}{
\begin{tabular}{lcccccccc}
\toprule
\multirow{2}{*}{\textbf{Benchmark}} & \multirow{2}{*}{\textbf{Tasks}} & \multicolumn{5}{c}{\textbf{Evaluation Category}} & \multirow{2}{*}{\textbf{Avg. Length}} & \multirow{2}{*}{\textbf{MLLM Eval.}} \\ \cline{3-7}
            &    & T2I & T2Is & I2I & Is2I & I(s)2Is &  \\ \midrule \midrule
PaintSkills \citep{cho2023dall} & 3  & \ding{51} & \ding{55} & \ding{55} & \ding{55} & \ding{55} & 10.44 & \ding{55} \\
GenEval \citep{ghosh2024geneval}    & 6  & \ding{51} & \ding{55} & \ding{55} & \ding{55} & \ding{55} & 7.61  & \ding{55} \\
EmuEdit \citep{sheynin2024emu}    & 12 & \ding{55} & \ding{55} & \ding{51} & \ding{55} & \ding{55} & 8.57  & \ding{55} \\
TIFA \citep{hu2023tifa}       & 12 & \ding{51} & \ding{55} & \ding{55} & \ding{55} & \ding{55} & 10.46 & GPT \citep{brown2020language} \\
DreamBooth \citep{ruiz2023dreambooth} & 30 & \ding{55} & \ding{55} & \ding{51} & \ding{55} & \ding{55} & 7.58  & \ding{55} \\
ImagenHub \citep{ku2023imagenhub} & 7  & \ding{51} & \ding{55} & \ding{51} & \ding{55} & \ding{55} & 8.61  & \ding{55} \\ 
\rowcolor{gray!20}
\BenchName  & \textbf{100} & \ding{51} & \ding{51} & \ding{51} & \ding{51} & \ding{51} & \textbf{138.68} & Gemini \citep{team2023gemini} \\ \bottomrule
\end{tabular}}
\label{tab:datasets}
\end{table}

The development of benchmarks typically keeps pace with advancements in text-to-image (T2I) synthesis \citep{ramesh2021zero,ramesh2022hierarchical,esser2021taming,rombach2022high,saharia2022photorealistic,betker2023improving,podell2023sdxl,esser2024scaling,baldridge2024imagen,chen2023pixart,blackforestlabs_flux_2024}. DrawBench was initially introduced by Imagen \citep{saharia2022photorealistic}, followed by DALL-EVAL \citep{cho2023dall}, which proposed PaintSkills to evaluate visual reasoning and social bias capabilities. Recently, an increasing number of benchmarks have emerged \citep{petsiuk2022human, bakr2023hrs, huang2023t2i, lee2024holistic, ku2023imagenhub}. While these benchmarks primarily focus on assessing image quality and alignment, DEsignBench \citep{lin2023designbench} shares similarities with our approach by emphasizing scenarios within authentic design contexts. Differently, \BenchName~extends beyond this by addressing additional open challenges in professional creativity.  Furthermore, \BenchName~is more closely aligned with the latest model capabilities, enabling the use of multi-image and complex instructions as diverse forms of guidance. As shown in \cref{tab:datasets}, \BenchName~greatly expand the prompt length and adde two evaluation dimensions: multi-image input and multi-image generation. This allows for the generation of both single and multiple images, thereby better mimicking the workflows of professional designers. 


%% file: sec/3_bench.tex
\section{\BenchName}
\label{versa}

We present \BenchName, a comprehensive image generation benchmark that encompasses a wide range of professional tasks in image generation and rigorously challenges models across multiple dimensions of capability. In this section, we first introduce \BenchName’s data collection over four different model level in \cref{sec:collection}, as well as the manual annotation pipeline in \cref{sec:annotation}, followed by a detailed explanation of both human evaluation and automated assessment methodologies, along with the specific metrics used, in \cref{sec:eval}.

\subsection{Data Collection}
\label{sec:collection}


To ensure that the tasks evaluated in \BenchName~closely mirror real-world and professional scenarios, we source all task directives and data from the internet and professional designers. Leveraging the knowledge and capabilities of GPT-4o \citep{gpt4o}, we clearly define and classify the tasks, creating task variants aligned with existing model capabilities. The tasks in \BenchName~are categorized as follows:

\begin{figure*}[!t]
\begin{center}
\includegraphics[width=1\linewidth]{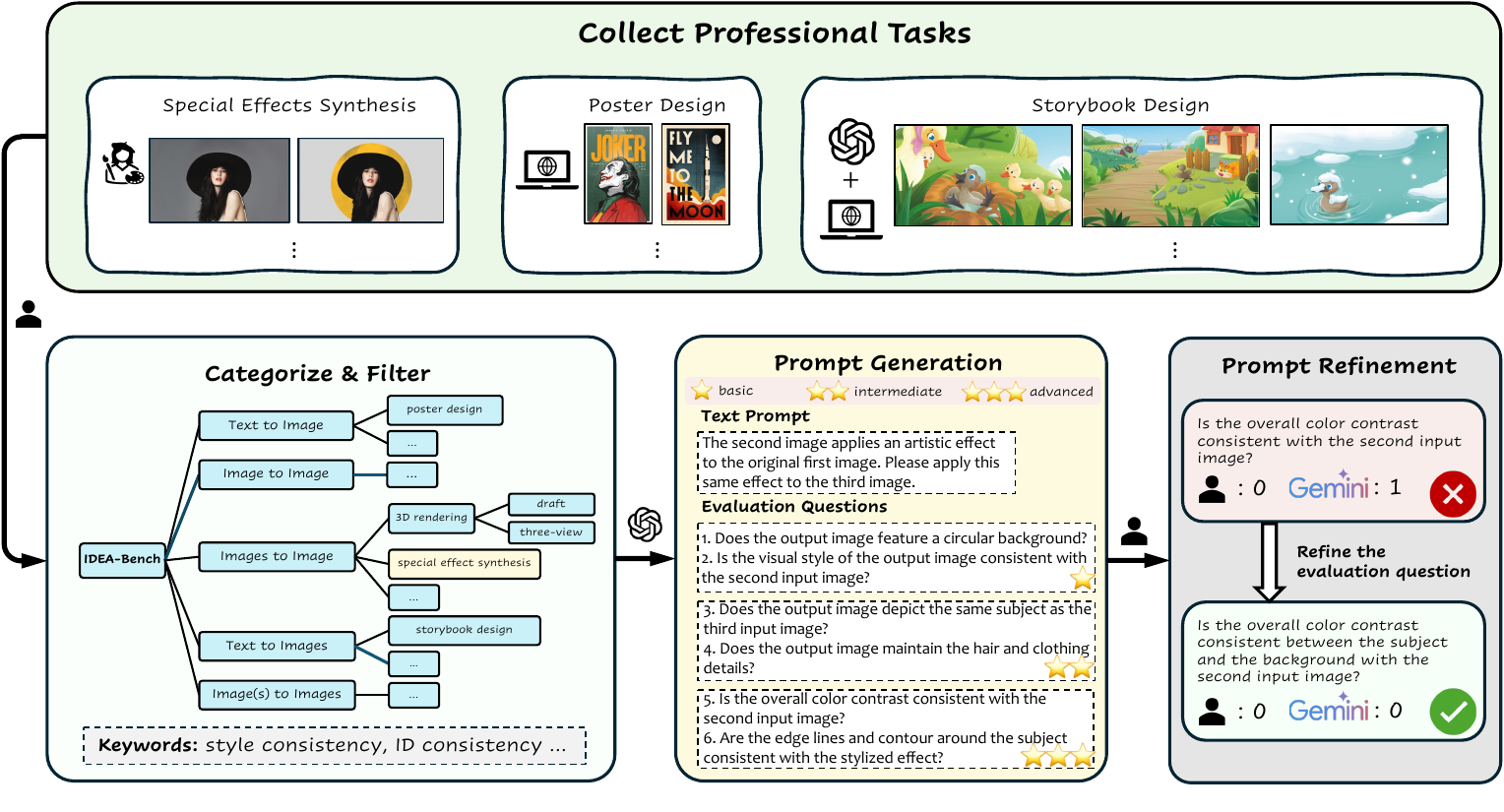}
\caption{\textbf{Dataset construction process of \BenchName.} We categorize the task data from professional design websites and designers based on generative model capabilities and assign capability keywords to each category. For each specific task, we design image generation prompts and hierarchical evaluation questions. Evaluators then refine these evaluation questions on a representative subset.} 
\label{fig:pipeline}
\end{center}
\end{figure*}

\paragraph{Text-only Guided Image Generation} 


While typical T2I users provide brief descriptions and rely on the model’s discretion, professional users meticulously control every visual detail, often iterating multiple prompts to achieve the desired outcome. Current T2I models \citep{ramesh2021zero,ramesh2022hierarchical,esser2021taming,rombach2022high,saharia2022photorealistic,betker2023improving,podell2023sdxl,esser2024scaling,baldridge2024imagen,chen2023pixart,blackforestlabs_flux_2024} struggle with complex prompts that include numerous visual concepts and lengthy text. To address this, \BenchName~includes 11 tasks designed to evaluate models’ ability to handle extensive visual components and incorporate text within images, essential for tasks like \textit{poster generation} and \textit{business card generation}.

\paragraph{Image / Multi-image Guided Image Generation}


Unlike ImagenHub \citep{ku2023imagenhub}, which categorizes conditional image generation tasks under specific definitions, we unify these into \textbf{image-to-image} and \textbf{images-to-image}. This consolidation promotes the development of models with enhanced task unification and generalization, similar to advancements in LLMs. \BenchName~includes numerous design tasks guided by both image and text inputs in real-world contexts, such as \textit{brand merchandise generation}, \textit{package rendering}, and \textit{image retouching}. With the advent of models like GDT \citep{huang2024group}, Emu2 \citep{sun2024generative}, and OmniGen \citep{xiao2024omnigen}, we anticipate future models will exhibit strong multimodal understanding and generative capabilities.

\paragraph{Text / (Multi-)image Guided Multi-image Generation}


Distinct from previous benchmarks \citep{cho2023dall, ghosh2024geneval, sheynin2024emu, hu2023tifa, ruiz2023dreambooth, ku2023imagenhub}, \BenchName~introduces tasks that require generating multiple related images simultaneously. These tasks cater to professional demands such as creating series images around a subject, multi-view image generation, and illustrating storybooks. Unlike video generation \citep{zhao2024real, yang2024cogvideox}, these image tasks involve significant visual differences while maintaining specific correlations. We further subdivide multi-image generation into \textbf{text-to-images} and \textbf{image(s)-to-images} categories to address models’ limitations in handling diverse input modalities, thereby testing their ability to maintain detail, style, and content consistency.

\subsection{Data Annotation}
\label{sec:annotation}

\paragraph{Prompt Generation}

After collecting all tasks and corresponding images, annotators provide clear task definitions for each specific case, which, along with input images, are fed to GPT-4o \citep{gpt4o} to generate image prompts tailored to each case. To ensure consistent evaluation across tasks, we also use GPT-4o to create a standardized set of six evaluation questions per task, ranging from basic to advanced difficulty levels. Specifically, recognizing the inherent subjectivity in both MLLM-based and human evaluations, we adopt a binary scoring system (0 or 1) for each question. This approach draws from methodologies in recent multimodal benchmarks \citep{bai2023touchstone, liu2025mmbench, li2023seed, zhang2024mme}, where multiple-choice formats are used to mitigate interpretive bias.

To further enhance objectivity and reduce variability, we modify the typical multi-level scoring scale found in existing datasets and models, converting it into a set of six binary (true/false) questions. Each question is accompanied by clearly defined scoring criteria, with 0 indicating unmet standards and 1 representing success. For deeper granularity, questions 1-2 assess \textbf{the model’s understanding of basic task requirements}, questions 3-4 evaluate \textbf{the quality of task completion}, and questions 5-6 examine \textbf{the model’s attention to detail and aesthetic quality} in the generated images. This structured assessment methodology ensures a rigorous and objective evaluation framework, aligning with professional standards and reducing potential sources of instability in scoring.

\paragraph{Prompt Refinement}

However, despite our efforts to define evaluation questions as objectively as possible, multimodal large language models (MLLMs) \citep{li2024llavanext-strong, wang2024qwen2, gpt4o, team2023gemini} still struggle to match human evaluators on complex, real-world tasks, limited by current models’ gaps in image comprehension relative to human capabilities. To address this, we select a subset of tasks and refine the evaluation questions based on specific examples, ensuring that MLLMs can yield intuitively reasonable results.

One challenge in using MLLMs as evaluators is the sensitivity of models like GPT-4V/GPT-4o to the order of image presentation \citep{wang2022self, wu2024gpt, zhang2023gpt, zheng2023judging}, which particularly makes these models less reliable for consistent comparative evaluation. MLLMs also fail to identify the sequence of multiple input images, whether it is input separately for multiple images or input as a collage. To mitigate this, we manually fine-tune evaluation questions for each case within the task subset, ensuring alignment with human intuition.

For example, in the task of \textit{generating a children’s storybook with defined character images}, we refine each evaluation angle by randomly sampling pairs of input and output images to test consistency. A typical question might ask \textit{if the primary character in the first input image appears with consistent features in the third output image}. For testing, we use the Gemini 1.5 Pro \citep{team2023gemini} to score model outputs and compared these results against human ratings. In cases where discrepancies arise, annotators iteratively refine the language of the evaluation questions until the MLLM evaluations align with human judgment. This approach enhances the robustness of our benchmark, ensuring that evaluations reflect both task-specific accuracy and consistency across diverse visual requirements.

\begin{figure*}[!t]
\begin{center}
\includegraphics[width=1\linewidth]{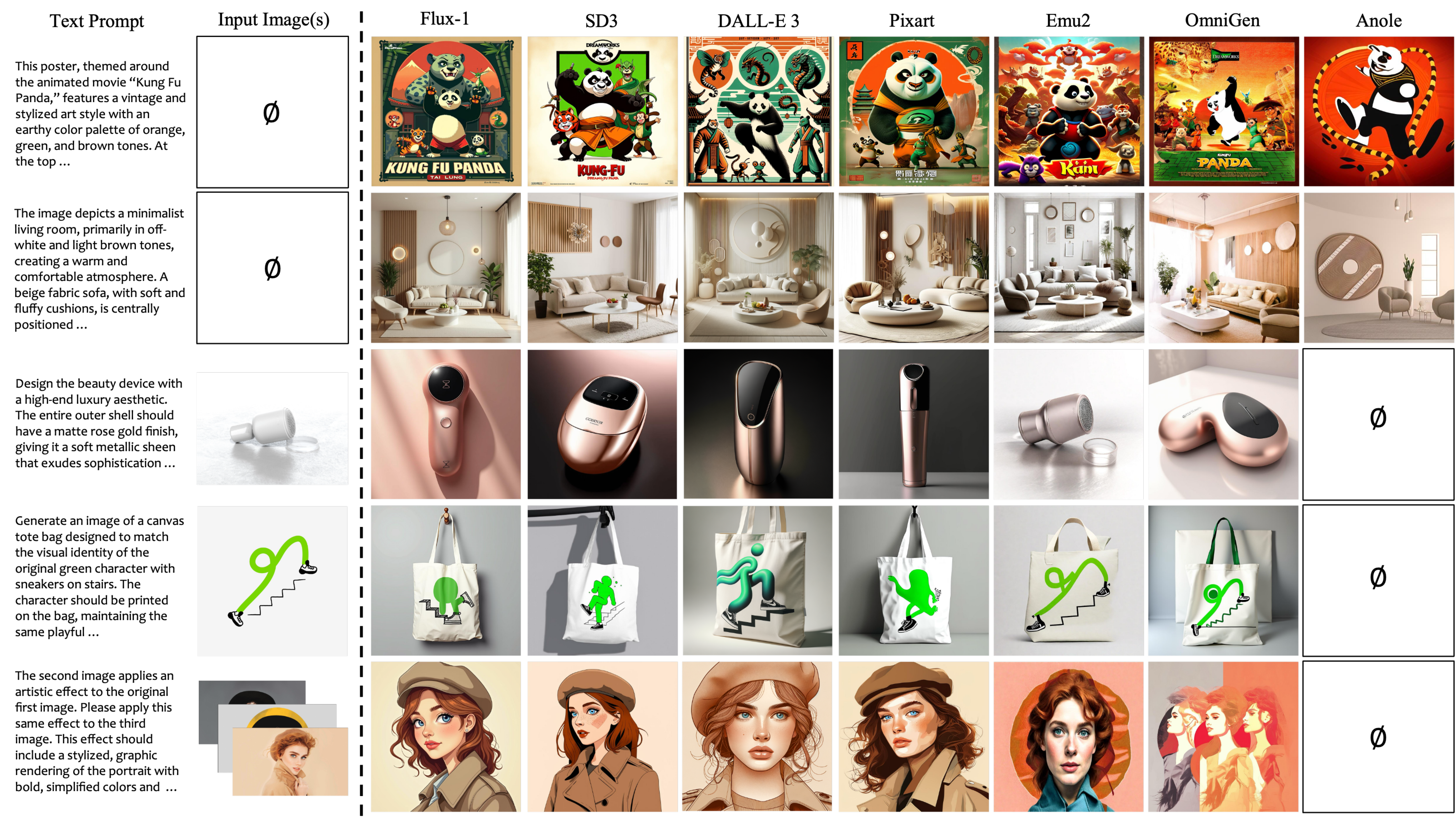}
\caption{\textbf{Visualization of model generated images on \BenchName. }We present the generation results of some models in representative task examples, including text-to-image, image-to-image, and images-to-image. In the case of no input image or no generated image, use "$\varnothing$" instead.} 
\label{fig:case-vis}
\end{center}
\vspace{-1em}
\end{figure*}

\subsection{Evaluation}
\label{sec:eval}

\subsubsection{Human Evaluation}
\label{sec:metric}

\paragraph{Evaluation Metric}

Each subtask in \BenchName~comprises 2 to 5 specific cases, with all cases within a subtask being evaluated using the same set of six binary (0 or 1) judgment questions. These evaluation questions are organized into three hierarchical levels, reflecting the priority of task completion over mere aesthetic quality. For each case, scoring progresses sequentially through these levels. If a model does not achieve a perfect score at a lower level, the scores for all higher levels are automatically set to 0. This hierarchical approach ensures that while a model may generate high-quality and visually appealing images, it must also fully meet the task requirements to receive a high overall score. This methodology aligns the evaluation criteria with human design standards, emphasizing the importance of fulfilling task objectives.








Formally, let the dataset be denoted as $\mathcal{D} = {\{\mathcal{D}_c | \{\mathcal{D}_c = {\{\mathcal{T}_t\}}^{T_c}_{t=1}\}}_{c=1}^{C}$. Each subtask $\mathcal{T}_t$ consists of 2 to 5 specific cases, all evaluated using the same set of six binary (0 or 1) judgment questions. The score for each subtask is calculated by averaging the binary scores of its cases. Subsequently, each major category $\mathcal{D}_c$ receives its score by averaging the scores of its subtasks and converting the result to a percentage. The overall score for the entire dataset $\mathcal{D}$ is then determined by averaging the scores of all $C$ major categories.

Importantly, if a model fails to complete a specific task, it is assigned a score of 0 for that task, ensuring that incomplete performances are accurately reflected in the overall evaluation. This hierarchical scoring framework ensures a precise and objective assessment of model performance, closely aligning with the rigorous standards expected by professional designers.

\subsubsection{Automated Evaluation}

Automated evaluation is performed on a subset of \BenchName~comprising 18 tasks, named \BenchName-mini. During the prompt refinement phase, annotators craft unique evaluation questions for each case, in contrast to the human evaluation setup where all cases within a subtask share the same 6 evaluation questions. This tailored approach allows for more precise and context-specific assessments. Scoring is conducted using Gemini 1.5 Pro \citep{team2023gemini}, adhering to the same hierarchical scoring methodology outlined in \cref{sec:metric}. To enhance result stability and reliability, each case is evaluated three times independently, and the final score is computed as the average of these three evaluations. This repetition mitigates potential variability in automated assessments and ensures consistent and robust scoring outcomes.

By integrating both human and automated evaluation methods, \BenchName~provides a comprehensive framework for assessing the multifaceted capabilities of generative models. This dual approach not only leverages the nuanced judgment of human evaluators but also benefits from the scalability and consistency of automated tools, thereby offering a balanced and thorough evaluation of model performance.

%% file: sec/4_eval.tex
\begin{table}[t]
\centering
\caption{\textbf{Experimental results on all categories of \BenchName.} Each task category is averaged across all its subtasks, with the top-ranked model scores for each task type highlighted in \textbf{bold}. Task types that a model cannot support are marked with "--" and are treated as 0 points in the average score calculation. "$\dagger$" represents rephrasing using GPT-4o to adapt the model to all tasks. }
\resizebox{0.7\textwidth}{!}{
\begin{tabular}{lccccccc}
\toprule
\multirow{2}{*}{\textbf{Method}} & \multirow{2}{*}{\textbf{Param.}} & \multicolumn{5}{c}{\textbf{Scores on All Categories}}                                          & \multirow{2}{*}{\textbf{Avg. Score}} \\ \cline{3-7}
                        &                         & T2I & I2I & Is2I & T2Is & I(s)2Is \\ \midrule \midrule
\rowcolor{gray!20}
FLUX-1$\dagger$ \citep{flux1}             & 12B & \textbf{46.06} & 12.13 & 4.89 & 20.15 & \textbf{29.17} & \textbf{22.48} \\
DALL-E 3$\dagger$ \citep{ramesh2022hierarchical}            & 12B & 24.34 & 6.95 & 5.27 & 14.36 & 14.44 & 13.07 \\
SD3$\dagger$ \citep{esser2024scalingrectifiedflowtransformers} & 2B & 24.04 & 10.79 & 4.69 & 21.59 & 13.06 & 14.83 \\
Pixart$\dagger$ \citep{chen2023pixart}             & 0.6B & 14.44 & 7.75 & 3.48 & 17.46 & 21.39 & 12.90 \\
InstructPix2Pix \citep{brooks2023instructpix2pix}    & 1B & -- & 17.58 & -- & -- & -- & 3.52 \\
MagicBrush \citep{zhang2024magicbrush}         & 1B & -- & \textbf{19.07} & -- & -- & -- & 3.81 \\
Anole \citep{chern2024anole}              & 7B & 0.00 & 0.64 & 0.00 & 1.74 & 0.00 & 0.48 \\
Emu2 \citep{sun2024generative}               & 37B & 17.98 & 7.05 & \textbf{8.98} & -- & -- & 6.81 \\
OmniGen \citep{xiao2024omnigen}            & 3.8B & 21.41 & 8.17 & 2.77 & -- & -- & 6.47 \\ \midrule
Emu2$\dagger$ \citep{sun2024generative}               & 37B & 17.98 & 7.05 & \textbf{8.98} & 15.53 & 12.78 & 12.46 \\
OmniGen$\dagger$ \citep{xiao2024omnigen}            & 3.8B & 21.41 & 8.17 & 2.77 & \textbf{23.52} & 21.39 & 15.45 \\ \bottomrule
\end{tabular}}
\label{tab:all}
\end{table}

\section{Experiments}
\label{expe}

\subsection{Experimental Details}

\subsubsection{Prompts Rephrasing}
\label{sec:rephrase}

Currently, only a few models \citep{yang2024seed, wu24next, wang2024emu3, chern2024anole} are capable of generating multiple related images simultaneously, makes it difficult to analyse a model's ability to generate multiple images potentially. However, as DALL-E 3 \citep{ramesh2022hierarchical} could perform multi-image generation owing to its MLLM GPT-4o \citep{gpt4o}, which could summarize multimodal input and then generate the corresponding prompt for each image. To empower other leading T2I models \citep{flux1, esser2024scaling, chen2023pixart}, we feed all input images and text prompt to GPT-4o and rephrase prompts for each image to be generated. By combining MLLM's multimodal understanding ability, we can approximately achieve the multi-image generation tasks even with basic T2I models. 

\subsubsection{Model Reimplementation}

We evaluate the capabilities of four categories of models in our study. For leading text-to-image (T2I) models, we select FLUX-1 \citep{flux1}, Stable Diffusion 3 (SD3) \citep{esser2024scaling}, Pixart \citep{chen2023pixart}, and DALL-E 3 \citep{ramesh2022hierarchical}. For image editing models, we choose InstructPix2Pix \citep{brooks2023instructpix2pix} and MagicBrush \cite{zhang2024magicbrush} To assess models with unified generation capabilities, we select Emu2 \citep{sun2024generative} and OmniGen \citep{xiao2024omnigen}. Additionally, for models that support interleaved text and image generation, we include Anole \citep{chern2024anole}. Notably, among the selected models, Anole \citep{chern2024anole} is the only one capable of handling an unrestricted number of input and output images. We endeavor to utilize the officially released model parameters and inference settings wherever possible; further details are provided in \cref{sec:config}. 

\begin{table*}[!t]
\centering
\caption{\textbf{Experimental results on Text-to-Image}. Each task category is averaged across all its subtasks, with the top-ranked model scores for each subtask highlighted in \textbf{bold}.}
\resizebox{1\textwidth}{!}{
\begin{tabular}{lcccccccccccc}
\toprule
\multirow{2}{*}{\textbf{Method}} &
  \multicolumn{11}{c}{\textbf{Subtasks Score}} &
  \multirow{2}{*}{\textbf{Avg. Score}} \\ \cline{2-12}
   &
  Arch. Style &
  Bus. Card &
  Game UI &
  Inf. Chart &
  Int.  &
  Paint. &
  Sculp. &
  Ticket &
  Land. &
  LOGO &
  Poster &
   \\ \midrule \midrule
\rowcolor{gray!20}
FLUX-1 \citep{flux1} & \textbf{100.00} & \textbf{38.89} & \textbf{5.56} & 0.00 & \textbf{66.67} & 61.11 & \textbf{16.67} & \textbf{16.67} & \textbf{83.33} & \textbf{61.11} & \textbf{56.67} & \textbf{46.06} \\
DALL-E 3 \citep{ramesh2022hierarchical} & 22.22 & 0.00 & 0.00 & 0.00 & 11.11 & \textbf{100.00} & 11.11 & 0.00 & 38.89 & \textbf{61.11} & 23.33 & 24.34 \\
SD3 \citep{esser2024scalingrectifiedflowtransformers} & 38.89 & 0.00 & \textbf{5.56} & 0.00 & 50.00 & 5.56 & \textbf{16.67} & 0.00 & 50.00 & \textbf{61.11} & 36.67 & 24.04 \\
Pixart \citep{chen2023pixart} & 5.56 & 0.00 & 0.00 & 0.00 & 22.22 & 33.33 & \textbf{16.67} & 0.00 & 22.22 & 22.22 & 36.67 & 14.44 \\
Anole \citep{chern2024anole} & 0.00 & 0.00 & 0.00 & 0.00 & 0.00 & 0.00 & 0.00 & 0.00 & 0.00 & 0.00 & 0.00 & 0.00 \\
Emu2 \citep{sun2024generative} & 22.22 & 0.00 & 0.00 & 0.00 & 38.89 & 38.89 & \textbf{16.67} & 0.00 & 44.44 & 16.67 & 20.00 & 17.98 \\
OmniGen \citep{xiao2024omnigen} & 38.89 & 0.00 & 0.00 & 0.00 & 27.78 & 50.00 & \textbf{16.67} & 0.00 & 38.89 & 33.33 & 30.00 & 21.41 \\ \bottomrule
\end{tabular}}
\label{tab:t2i-score}
\end{table*}

\begin{table}[!t]
\caption{\textbf{Experimental results on Image-to-Image.} Each task category is averaged across all its subtasks, with the top-ranked model scores for each subtask highlighted in \textbf{bold}. "$\dagger$" represents the use of MLLM for prompt rephrasing.}
\resizebox{\textwidth}{!}{
\begin{tabular}{lcccccccccccccc}
\toprule
\multirow{2}{*}{\textbf{Method}} &
  \multicolumn{13}{c}{\textbf{Subtasks Score}} &
  \multirow{2}{*}{\textbf{Avg. Score}} \\ \cline{2-14}
   &
  Anim. Attr. &
  Coup. Icon &
  ID Photo &
  Blur &
  Compl. &
  Ret. &
  Str. &
  Light. &
  Enl. &
  Obj. Edit &
  Pack. Rend. &
  Style Edit &
  Brand Merch. & 
   \\ \midrule \midrule
FLUX-1$\dagger$ \citep{flux1} & 0.00 & 16.67 & 0.00 & 2.78 & 4.17 & 0.00 & 12.50 & 0.00 & \textbf{100.00} & 6.25 & 0.00 & 4.17 & 11.11 & 12.13 \\
DALL-E 3$\dagger$ \citep{ramesh2022hierarchical} & 5.56 & 11.11 & 0.00 & 2.78 & 0.00 & 0.00 & 4.17 & 0.00 & 38.89 & 4.17 & 8.33 & 4.17 & 11.11 & 6.95 \\
SD3$\dagger$ \citep{esser2024scalingrectifiedflowtransformers} & 8.33 & 27.78 & 0.00 & 2.78 & 4.17 & 0.00 & 8.33 & 0.00 & 55.56 & 4.17 & 8.33 & 4.17 & 16.67 & 10.79 \\
Pixart$\dagger$ \citep{chen2023pixart} & 5.56 & \textbf{55.56} & 4.17 & 2.78 & 4.17 & 0.00 & 0.00 & 0.00 & 11.11 & 4.17 & 0.00 & 2.08 & 11.11 & 7.75 \\
InstructPix2Pix \citep{brooks2023instructpix2pix} & \textbf{44.44} & 16.67 & \textbf{16.67} & 13.89 & \textbf{25.00} & \textbf{54.17} & 0.00 & 0.00 & 0.00 & 10.42 & \textbf{25.00} & 16.67 & 5.56 & 17.58 \\
\rowcolor{gray!20}
MagicBrush \citep{zhang2024magicbrush} & 33.33 & 16.67 & 12.50 & \textbf{25.00} & \textbf{25.00} & 41.67 & 0.00 & 0.00 & 0.00 & \textbf{35.42} & 8.33 & \textbf{33.33} & 16.67 & \textbf{19.07} \\
Anole \citep{chern2024anole} & 0.00 & 0.00 & 0.00 & 0.00 & 8.33 & 0.00 & 0.00 & 0.00 & 0.00 & 0.00 & 0.00 & 0.00 & 0.00 & 0.64 \\
Emu2 \citep{sun2024generative} & 11.11 & 0.00 & 4.17 & 5.56 & 0.00 & 0.00 & \textbf{20.83} & 0.00 & 16.67 & 4.17 & 8.33 & 4.17 & 16.67 & 7.05 \\
OmniGen \citep{xiao2024omnigen} & 16.67 & 0.00 & 4.17 & 5.56 & 0.00 & 4.17 & 4.17 & 0.00 & 0.00 & 16.67 & 0.00 & 10.42 & \textbf{44.44} & 8.17 \\ \bottomrule
\end{tabular}}
\label{tab:i2i-score}
\end{table}

\subsection{Results on \BenchName}

\paragraph{Qualitative Analysis}

\Cref{fig:case-vis} showcases the generated images across 7 models. For ease of comparison, we select several cases from three categories: \textbf{text-to-image}, \textbf{image-to-image,} and \textbf{images-to-image}. In fundamental \textbf{text-to-image} tasks, FLUX-1 \citep{flux1} demonstrates a clear advantage over other models, effectively comprehending the visual elements and professional-level requirements outlined in the prompts. Regarding image editing tasks, models that are not specifically designed for image editing struggle to preserve essential information from the original image. Emu2 \citep{sun2024generative} is the only model that roughly restores the shapes and compositions of objects from the original image. OmniGen \citep{xiao2024omnigen} excels in maintaining the identity of objects. However, for more complex tasks such as special effect synthesis, no current model sufficiently understands the intricate demands of these professional-level requirements.

We first present the ranking of all models based on their scores across the entire dataset, as detailed in \cref{tab:all}. Leveraging GPT-4o’s \citep{gpt4o} multimodal understanding and task translation capabilities, FLUX-1 \citep{flux1} achieve the highest score of 22.46, significantly outperforming other models. Specialized image editing models \citep{brooks2023instructpix2pix, zhang2024magicbrush} lead in the \textbf{image-to-image} tasks, while the general-purpose generation model Emu2 \citep{sun2024generative} excels in the \textbf{images-to-image} tasks. In contrast, Anole \citep{chern2024anole}, which is capable of generating multiple images simultaneously, do not achieve an ideal final score. Furthermore, no model surpass a score of 10 without utilizing the comprehension capabilities of MLLMs, indicating that general-purpose image generation models still face substantial challenges in achieving professional-grade performance. This underscores the need for comprehensive evaluation standards to guide the development of more specialized and capable generative models. 

To further quantify the impact of MLLM \citep{gpt4o} assistance on model capabilities within the benchmark, we proceed \textbf{text-to-images} and \textbf{image(s)-to-images} tasks that not supported by Emu2 \citep{sun2024generative} and OmniGen \citep{xiao2024omnigen} using GPT-4o-rephrased prompts. The results are supplemented at the bottom of \cref{tab:all}. Despite enhancing Emu2 and OmniGen’s capabilities with MLLMs, FLUX-1 \citep{flux1} remains the top performer, as shown in \cref{tab:all-mllm}. Notably, FLUX-1 even surpasses other models that support image input in \textbf{image-to-image} tasks. The advantage of T2I models \citep{flux1, ramesh2022hierarchical, esser2024scaling, chen2023pixart} in this experiment lies in their ability to leverage MLLMs to understand different tasks. \BenchName’s task definitions are highly specialized, making it difficult for other models to comprehend these tasks without relying on MLLM. Universal generative models also have difficult ensuring the quality of generated images, resulting in lower scores finally. Overall, to achieve high scores across all benchmark tasks, a model must possess both multimodal input-output capabilities and robust MLLM-level multimodal understanding. The following section provides a detailed analysis of each category.

\subsubsection{Results on Text-to-Image}

\begin{table}[t]
\centering
\caption{\textbf{Experimental results on Images-to-Image.} Each task category is averaged across all its subtasks, with the top-ranked model scores for each subtask highlighted in \textbf{bold}. "$\dagger$" represents the use of MLLM for prompt rephrasing.}
\resizebox{1\textwidth}{!}{
\begin{tabular}{lccccccccc}
\toprule
\multirow{2}{*}{\textbf{Method}} &
  \multicolumn{8}{c}{\textbf{Subtasks Score}} &
  \multirow{2}{*}{\textbf{Avg. Score}} \\ \cline{2-9}
   &
  Human Attr. &
  Image Trans. &
  Pan. &
  Text Edit. &
  3D. Effect &
  Image Blend. &
  Spec. Effect &
  Real and Anime &
   \\ \midrule \midrule
FLUX-1$\dagger$ \citep{flux1} & 3.97 & \textbf{6.67} & 0.00 & \textbf{10.42} & 0.00 & \textbf{12.50} & 0.00 & 5.56 & 4.89 \\
DALL-E 3$\dagger$ \citep{ramesh2022hierarchical} & 4.76 & 3.33 & 8.33 & 6.25 & 0.00 & 8.33 & 5.56 & 5.56 & 5.27 \\
SD3$\dagger$ \citep{esser2024scalingrectifiedflowtransformers} & 4.76 & 5.00 & 0.00 & 8.33 & 0.00 & 8.33 & 5.56 & 5.56 & 4.69 \\
Pixart$\dagger$ \citep{chen2023pixart} & 4.76 & 5.00 & 8.33 & 0.00 & 0.00 & 4.17 & 0.00 & 5.56 & 3.48 \\
Anole \citep{chern2024anole} & 0.00 & 0.00 & 0.00 & 0.00 & 0.00 & 0.00 & 0.00 & 0.00 & 0.00 \\
\rowcolor{gray!20}
Emu2 \citep{sun2024generative} & \textbf{6.35} & 0.00 & \textbf{50.00} & 2.08 &  0.00 & 4.17 & 0.00 & \textbf{9.26} & \textbf{8.98} \\
OmniGen \citep{xiao2024omnigen} & 1.59 & 0.00 & 0.00 & 2.08 & 0.00 & 0.00 & \textbf{11.11} & 7.41 & 2.77 \\ \bottomrule
\end{tabular}}
\label{tab:is2i-score}
\end{table}

\begin{table*}[t]
\caption{\textbf{Experimental results on Text-to-Images.} Each task category is averaged across all its subtasks, with the top-ranked model scores for each subtask highlighted in \textbf{bold}. "$\dagger$" represents the use of MLLM for prompt rephrasing.}
\resizebox{\textwidth}{!}{
\begin{tabular}{lcccccccccccc}
\toprule
\multirow{2}{*}{\textbf{Method}} &
  \multicolumn{11}{c}{\textbf{Subtasks Score}} &
  \multirow{2}{*}{\textbf{Avg. Score}} \\ \cline{2-12}
   &
  Anim. Grow. &
  Child. Book &
  Draw. Proc. &
  Hist. Narr. &
  Movie Shots &
  Phys. Laws &
  Plant Grow. &
  Style Group &
  Conc. Vis. &
  Coup. Icon &
  Dyn. Char. &
   \\ \midrule \midrule
FLUX-1$\dagger$ \citep{flux1} & 25.00 & \textbf{4.17} & 8.33 & \textbf{70.83} & 4.17 & 4.17 & 8.33 & 7.08 & \textbf{16.67} & \textbf{50.00} & \textbf{22.92} & 20.15 \\
DALL-E 3$\dagger$ \citep{ramesh2022hierarchical} & 0.00 & \textbf{4.17} & 0.00 & 4.17 & 4.17 & 0.00 & 41.67 & \textbf{32.92} & 12.50 & 41.67 & 16.67 & 14.36 \\
\rowcolor{gray!20}
SD3$\dagger$ \citep{esser2024scalingrectifiedflowtransformers} & 8.33 & \textbf{4.17} & \textbf{25.00} & 37.50 & \textbf{12.50} & \textbf{8.33} & \textbf{45.83} & 18.75 & \textbf{16.67} & 41.67 & 18.75 & \textbf{21.59} \\
Pixart$\dagger$ \citep{chen2023pixart} & \textbf{37.50} & \textbf{4.17} & 4.17 & 16.67 & 0.00 & 0.00 & 29.17 & 21.25 & \textbf{16.67} & 45.83 & 16.67 & 17.46 \\
Anole \citep{chern2024anole} & 0.00 & 0.00 & 0.00 & 0.00 & 0.00 & 0.00 & 0.00 & 0.42 & 8.33 & 8.33 & 2.08 & 1.74 \\
\bottomrule
\end{tabular}}
\label{tab:t2is-score}
\end{table*}

\begin{table*}[t]
\caption{\textbf{Experimental results on Image(s)-to-Images.} Each task category is averaged across all its subtasks, with the top-ranked model scores for each subtask highlighted in \textbf{bold}. "$\dagger$" represents the use of MLLM for prompt rephrasing.}
\resizebox{\textwidth}{!}{
\begin{tabular}{lccccccccc}
\toprule
\multirow{2}{*}{\textbf{Method}} &
  \multicolumn{8}{c}{\textbf{Subtasks Score}} &
  \multirow{2}{*}{\textbf{Avg. Score}} \\ \cline{2-9}
   &
  Anim. Grow. &
  Creat. Trans. &
  Layer Decomp. &
  Light. Effect & 
  Movie Shots &
  Multi-app. &
  Multi-dec. &
  Multi-view Trans. &
   \\ \midrule \midrule
\rowcolor{gray!20}
FLUX-1$\dagger$ \citep{flux1} & 25.00 & \textbf{66.67} & \textbf{75.00} & 8.33 & 0.00 & 0.00 & 16.67 & \textbf{100.00} & \textbf{29.17} \\
DALL-E 3$\dagger$ \citep{ramesh2022hierarchical} & 25.00 & 25.00 & 8.33 & 0.00 & \textbf{4.17} & 0.00 & 16.67 & 8.33 & 14.44 \\
SD3$\dagger$ \citep{esser2024scalingrectifiedflowtransformers} & \textbf{33.33} & 50.00 & 0.00 & 0.00 & 0.00 & 0.00 & 8.33 & 0.00 & 13.06 \\
Pixart$\dagger$ \citep{chen2023pixart}  & \textbf{33.33} & 50.00 & 0.00 & \textbf{58.33} & 0.00 & 0.00 & \textbf{58.33} & 0.00 & 21.39 \\
Anole \citep{chern2024anole} & 0.00 & 0.00 & 0.00 & 0.00 & 0.00 & 0.00 & 0.00 & 0.00 & 0.00 \\
\bottomrule
\end{tabular}}
\label{tab:is2is-score}
\end{table*}

\begin{table*}[t]
\centering
\caption{\textbf{Experimental results on the subset \BenchName-mini.} Each task category is averaged across all its subtasks, with the top-ranked model scores for each subtask highlighted in \textbf{bold}. "G." represents evaluate using Gemini 1.5 pro \citep{team2023gemini}, "H." represents human evaluation. "$\dagger$" represents the use of MLLM for prompt rephrasing.}
\resizebox{1\textwidth}{!}{
\begin{tabular}{lccccccccccccc}
\toprule
\multirow{3}{*}{\textbf{Method}} &
  \multirow{3}{*}{\textbf{CLIPScore}} &
  \multicolumn{10}{c}{\textbf{Evaluation Category}} &
  \multicolumn{2}{c}{\multirow{2}{*}{\textbf{Avg. Score}}} \\ \cline{3-12}
 &
   &
  \multicolumn{2}{c}{T2I} &
  \multicolumn{2}{c}{I2I} &
  \multicolumn{2}{c}{Is2I} &
  \multicolumn{2}{c}{T2Is} &
  \multicolumn{2}{c}{I(s)2Is} &
  \multicolumn{2}{c}{} \\ \cline{3-14} 
                   &  & G. & H. & G. & H. & G. & H. & G. & H. & G. & H. & G. & H. \\ \midrule \midrule
\rowcolor{gray!20}
FLUX-1$\dagger$ \citep{flux1}             & 0.3432 & 83.33 & 67.04 & 28.71 & 5.56 & 4.86 & 6.25 & 32.29 & 28.13 & 37.50 & 38.33 & \textbf{37.48} & \textbf{29.06} \\
DALL-E 3$\dagger$ \citep{ramesh2022hierarchical}            & 0.3462 & 55.56 & 41.11 & 27.78 & 5.56 & 20.37 & 5.56 & 29.17 & 21.88 & 29.17 & 33.33 & 32.41 & 21.49 \\
SD3$\dagger$ \citep{esser2024scalingrectifiedflowtransformers} & \textbf{0.3483} & 56.30 & 39.26 & 23.61 & 8.34 & 5.55 & 5.56 & 32.29 & 23.96 & 31.25 & 29.80 & 29.80 & 19.76 \\
Pixart$\dagger$ \citep{chen2023pixart}             & 0.3435 & 26.30 & 27.04 & 30.55 & 5.56 & 29.86 & 2.09 & 35.42 & 23.96 & 31.25 & 30.00 & 30.68 & 17.73 \\
InstructPix2Pix \citep{brooks2023instructpix2pix}    & 0.1194 & -- & -- & 27.78 & 2.78 & -- & -- & -- & -- & -- & -- & 5.56 & 0.56 \\
MagicBrush \citep{zhang2024magicbrush}         & 0.1212 & -- & -- & 9.72 & 8.34 & -- & -- & -- & -- & -- & -- & 1.94 & 1.67 \\
Emu2 \citep{sun2024generative}               & 0.1821 & 35.80 & 27.04 & 43.52 & 8.34 & 34.72 & 2.09 & -- & -- & -- & -- & 22.81 & 7.49 \\
OmniGen \citep{xiao2024omnigen}            & 0.1799 & 46.30 & 34.07 & 40.27 & 11.11 & 25.46 & 0.00 & -- & -- & -- & -- & 22.41 & 6.81 \\ \bottomrule 
\end{tabular}}
\label{tab:mini-score}
\end{table*}

All models’ results on \textbf{text-to-image} tasks are presented in \cref{tab:t2i-score}. Notably, the FLUX-1 \citep{flux1} model significantly outperforms the others, effectively accomplishing the task objectives in the majority of tasks. Based on the task settings and each model’s performance, we observe the following:

\begin{itemize}
    \item \textbf{Text Generation Challenges:} Text generation tasks are a common pain point for all models. Even though models can generate specific visual elements with high quality and understand style instructions specified in the prompts, they struggle to incorporate text into images as seamlessly as human designers. For example, in the 
    \textit{business card generation} task, only FLUX-1 \citep{flux1} received scores.

    \item \textbf{Performance of Basic \textit{vs.} General Models:} Basic text-to-image models generally demonstrate higher task completion rates compared to general models like Emu2 \citep{sun2024generative} and OmniGen \citep{xiao2024omnigen}. This is expected, as general models prioritize enhancing task generalization and multimodal information processing capabilities, which inevitably compromises some aspects of image generation quality.

    \item \textbf{Poor Performance on Information Charts:} All models perform poorly on the \textit{information chart generation} task, receiving a score of 0 from human evaluators. The primary reason is that generative models struggle to accurately transfer large amounts of text from the input to the output image, which is a basic ability to human designers. Additionally, many models do not support ultra-long text inputs.
\end{itemize}

\subsubsection{Results on Image-to-Image \& Images-to-Image}

The \textbf{image-to-image} tasks represent the strengths of image editing models. In \cref{tab:i2i-score}, MagicBrush \citep{zhang2024magicbrush} and InstructPix2Pix \citep{brooks2023instructpix2pix} achieve the first and second highest evaluation scores, respectively. The \textbf{images-to-image} tasks involve more complex and diverse guidance or conditions, with Emu2 \citep{sun2024generative} attaining the highest score of 8.98, as shwon in \cref{tab:is2i-score}. From these results, we derive the following key insights:

\begin{itemize}
    \item \textbf{Pros and Cons of Image Editing Models:} In local editing tasks such as \textit{image blur} and \textit{image retouching}, image editing models excel at preserving the original features of the image. However, these models are typically trained or fine-tuned on specific image editing datasets, which diminishes their ability to perform subject-driven tasks, such as \textit{branded merchandise generation}.
    
    \item \textbf{In-Context Learning Capabilities:} Emu2 \citep{sun2024generative} and OmniGen \citep{xiao2024omnigen} exhibit in-context learning abilities, as evidenced by OmniGen’s significantly higher scores in tasks like \textit{branded merchandise generation} and \textit{special effect synthesis}, achieving ID consistency between input and output images. Emu2 demonstrates the ability to comprehend and partially complete the \textit{image straighten} task, indicating that the model possesses an understanding of the semantic aspects of visual elements.

    \item \textbf{Challenges for Generative Models:} General-purpose models perform poorly in tasks such as \textit{light condition editing}. Although these tasks are frequently encountered by designers, the models require a certain level of understanding of real-world physical models, presenting a challenge for future generative models.
\end{itemize}

\subsubsection{Results on Text-to-Images \& Image(s)-to-Images}

\Cref{tab:t2is-score} and \cref{tab:is2is-score} present the evaluation results for \textbf{text-to-images} and select \textbf{image(s)-to-images} tasks, respectively. In both categories, SD3 \citep{esser2024scaling} and FLUX-1 \citep{flux1} achieve the highest scores. For tasks such as \textit{historical event generation}, MLLMs can effectively convey detailed style requirements during the prompt rephrasing phase, resulting in relatively consistent style outcomes. However, for more complex tasks like \textit{children’s book generation} that involve sophisticated capabilities such as consistency and preservation over ID and style, existing models struggle to perform adequately even when assisted by MLLMs. Nonetheless, generating multiple related images holds significant value; for example, the generated movie shots can serve as guiding conditions for multi-shot video generation models. We aim for the task types designed in our benchmark to align generative models with the capabilities of human designers, thereby further advancing toward general AI.

\subsubsection{Results on \BenchName-mini} 

\Cref{tab:mini-score} presents the evaluation results for a subset of 18 tasks within our benchmark, where we calculate CLIPScore \citep{hessel2021clipscore}, Gemini 1.5 Pro \citep{team2023gemini} scores, and human evaluation scores. As shown in \cref{tab:mini-score}, FLUX-1 \citep{flux1}, leveraging MLLM’s prompt rephrasing capabilities, achieve the highest scores in both MLLM-based and human evaluations within this subset. In contrast, when excluding the use of MLLMs, Emu2 \citep{sun2024generative} secure the top score. These experimental results are consistent with the overall benchmark dataset findings. Additionally, compared to CLIPScore, MLLM-based evaluations more closely align with human preferences in terms of professional evaluation data and assessment criteria. This suggests that as MLLM capabilities continue to evolve, utilizing MLLMs for image generation evaluation will become increasingly stable and reliable.

%% file: sec/5_conclusion.tex
\section{Conclusion}
\label{sec:conclusion}

We introduce \BenchName, a comprehensive benchmark designed to bridge the gap between current generative model capabilities and the stringent demands of professional-grade image design. \BenchName~encompasses 100 professional image generation tasks across five distinct categories, utilizing a detailed evaluation framework to ensure precise and objective assessments. Additionally, we build a representative subset IDEA-Bench-mini that facilitates automated evaluation using MLLMs. Our evaluations reveal significant gaps in existing models, particularly in handling intricate instructions and maintaining consistency in multi-image generation, underscoring the need for more advanced and versatile models. By aligning evaluation standards with the nuanced requirements of human designers, \BenchName~serves as a pivotal tool for guiding the development of generative models toward achieving professional-grade performance and advancing toward general artificial intelligence with autonomous and sophisticated visual generation capabilities. 

\newpage

%% file: sec/X_suppl.tex
\clearpage
\setcounter{page}{1}


\section{Implementation Details}

In this section, we detail the methods used for \BenchName~construction and experimental anlyses to ensure reproducibility. \Cref{sec:instruction} provides example instructions for utilizing GPT-4o \citep{gpt4o} in the construction of \BenchName~, while \cref{sec:config} outlines the experimental configurations.

\subsection{\BenchName~Construction Instruction}
\label{sec:instruction}

\paragraph{Instruction for prompt rephrasing}

\begin{figure*}[!t]
\centering
\begin{center}
\includegraphics[width=0.9\linewidth]{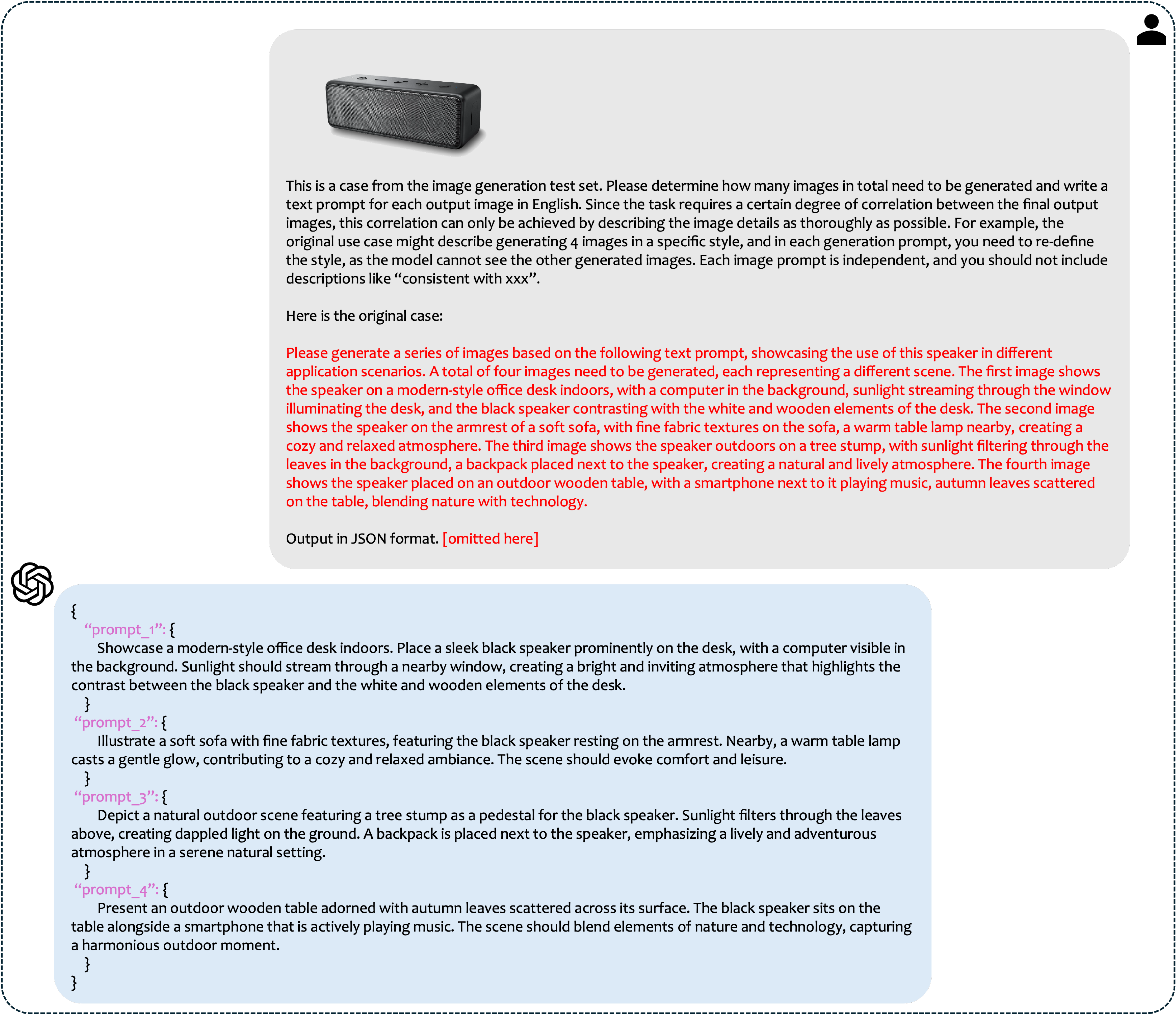}
\caption{\textbf{An instruction example for prompt rephrasing.} The red sections indicate customization for different subtasks. The JSON file format templates within the instructions are not shown.} 
\label{fig:gen-t2i}
\end{center}
\end{figure*}

As mentioned in \cref{sec:rephrase}, to closely align with real design scenarios, \BenchName~includes multi-image generation tasks that most existing models do not support. To thoroughly evaluate current generative models’ capabilities in these tasks, we utilize one of the most advanced MLLMs, GPT-4o \citep{gpt4o}, to rephrase multimodal inputs (which may include multiple images and complex long texts) into several text-to-image prompts. The specific rephrasing instruction is illustrated in \cref{fig:gen-t2i}. However, transforming tasks through rephrasing is merely a workaround, as text alone cannot capture all the details of the given images. Human designers have the ability to autonomously extract information from images and transform it into outputs in a freeform manner. We aim for \BenchName~to drive future generative models to acquire this capability.

\paragraph{Instruction for evaluation question construction}

\begin{figure*}[!t]
\centering
\begin{center}
\includegraphics[width=0.9\linewidth]{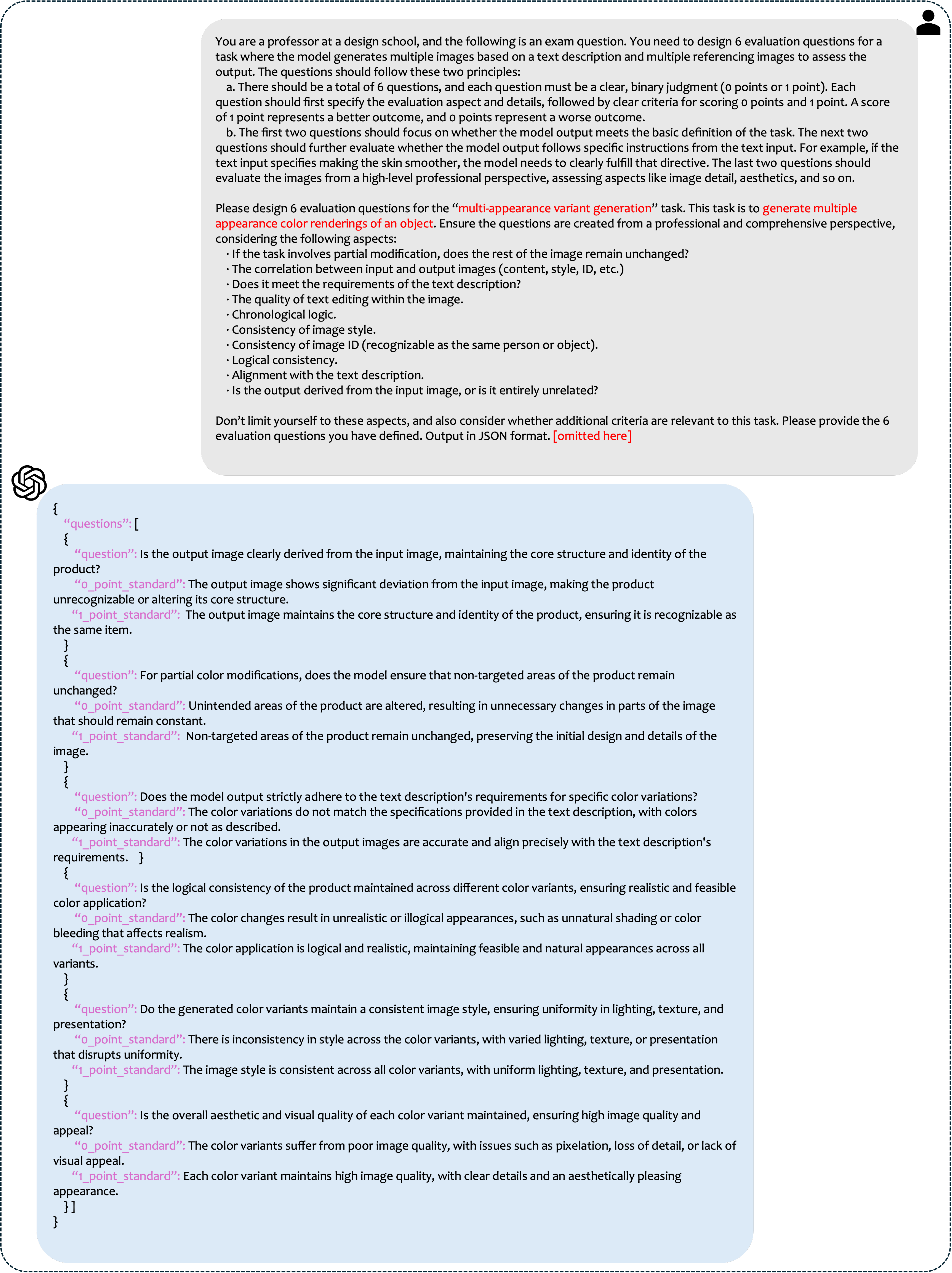}
\caption{\textbf{An instruction example for generating evaluation questions.} The red sections indicate customization for different subtasks. The JSON file format templates within the instructions are not shown.} 
\label{fig:gen-ques}
\end{center}
\end{figure*}

After collecting the task data, we generate evaluation questions in bulk by combining task keywords provided by human annotators with GPT-4o \citep{gpt4o}. \Cref{fig:gen-ques} illustrates an example of the instruction for generating evaluation questions for \textbf{image(s)-to-images} tasks. In \cref{fig:gen-ques}, the red sections indicate prompts that need to be customized for each specific task, while the JSON format templates are omitted. Within the fixed prompts, we first outline the basic requirements for the evaluation questions, such as multi-level standards, the exclusive use of objective judgment questions, and the convention that a score of 1 signifies a better result compared to 0. After incorporating the fundamental task definitions provided by annotators, the prompts also include frequently occurring evaluation capability keywords specific to multi-image generation tasks. This ensures that the evaluation questions defined by GPT-4o maintain a professional standard.

\subsection{Inference Configuration}
\label{sec:config}

\Cref{tab:config} details the configurations applied during inference for all models. To ensure fairness, all diffusion-based models employ 50 sampling steps (DALL-E 3 \citep{ramesh2022hierarchical} utilizes the official API and is therefore excluded from the statistics). Notably, Anole's visual decoder is not diffusion-based \citep{chern2024anole}; instead, it employs a diffusion-free, token-based architecture. We adhere to the text guidance scale and image guidance scale recommended by the official project codes, as illustrated in \cref{tab:config}.

\section{Statistical Analysis}

\Cref{fig:overview} visualizes the distribution of all subtasks across categories. In this section, we further conduct statistical analyses on the composition of the prompts and evaluation criteria of \BenchName.

\paragraph{Distribution of prompt length.}

\begin{figure*}[!t]
\centering
\begin{center}
\includegraphics[width=0.8\linewidth]{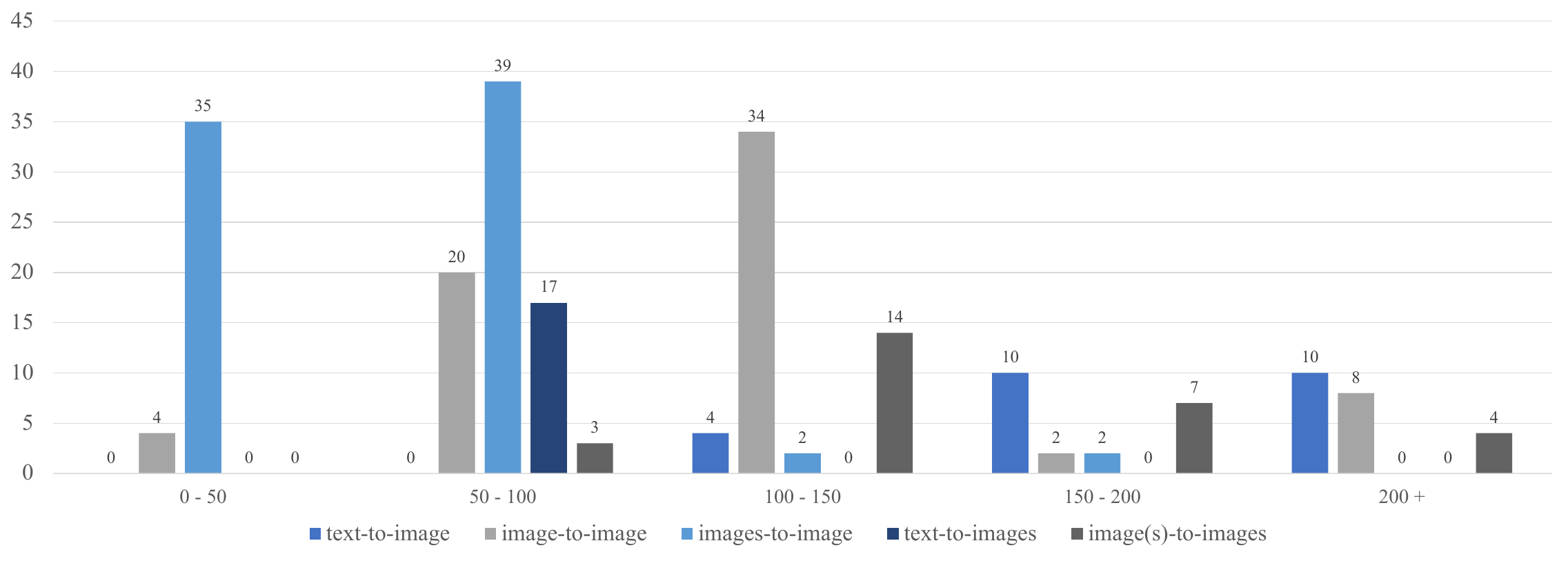}
\caption{\textbf{Statistics of prompt lengths for all tasks in \BenchName.} Each of the five task categories is represented by a distinct color. Prompt lengths are divided into five intervals, and the y-axis shows the number of tasks that fall within each interval.} 
\label{fig:prompt-length}
\end{center}
\end{figure*}

In \cref{fig:prompt-length}, we present the distribution of prompt lengths across the five task categories using histograms. According to the statistics in \cref{tab:datasets}, \BenchName’s prompts have an average length of approximately 139 words. Prompts shorter than the average are primarily found in the \textbf{image-to-image} and \textbf{images-to-image} tasks, as these tasks rely heavily on input images to guide the final generation, reducing the need for extensive textual descriptions. However, the prompt lengths for these two categories still significantly exceed those of other benchmarks \citep{cho2023dall, ghosh2024geneval, sheynin2024emu, hu2023tifa, ruiz2023dreambooth, ku2023imagenhub}. Additionally, both \textbf{text-to-image} and \textbf{image-to-images} tasks feature excessively long prompts, due to the requirements for complex and rich visual elements or detailed descriptions for multiple generated images.

\begin{table}[!t]
\centering
\caption{\textbf{Inference details of the models being tested.} "--" indicates either an API call or the absence of relevant parameters.}
\resizebox{0.8\textwidth}{!}{
\begin{tabular}{lccccc}
\toprule
\textbf{Method}    & \textbf{Param.} & \textbf{DiT based} & \textbf{Text Guid. Scale} & \textbf{Image Guid. Scale} & \textbf{Steps} \\ \midrule \midrule
FLUX-1~\citep{flux1} & 12B & \ding{51} & 3.5 & -- & 50 \\
DALL-E 3~\citep{ramesh2022hierarchical} & 12B & \ding{55} & -- & -- & -- \\
SD3~\citep{esser2024scaling} & 2B & \ding{51} & 7.0 & -- & 50 \\
Pixart~\citep{chen2023pixart} & 0.6B & \ding{51} & 7.0 & -- & 50 \\
InstructPix2Pix~\citep{brooks2023instructpix2pix} & 1B & \ding{55} & 7.5 & 1.5 & 50 \\
MagicBrush~\citep{zhang2024magicbrush} & 1B & \ding{55} & 7.5 & 1.5 & 50 \\
Emu2~\citep{sun2024generative} & 37B & \ding{55} & 3.0 & -- & 50 \\
OmniGen~\citep{xiao2024omnigen} & 3.8B & \ding{51} & 3.0 & 1.6 & 50 \\
Anole~\citep{chern2024anole} & 7B & -- & -- & -- & -- \\ \bottomrule
\end{tabular}}
\label{tab:config}
\end{table}

\paragraph{Distribution of evaluation ability.}

\begin{figure}[!t]
\centering
\begin{center}
\includegraphics[width=0.7\linewidth]{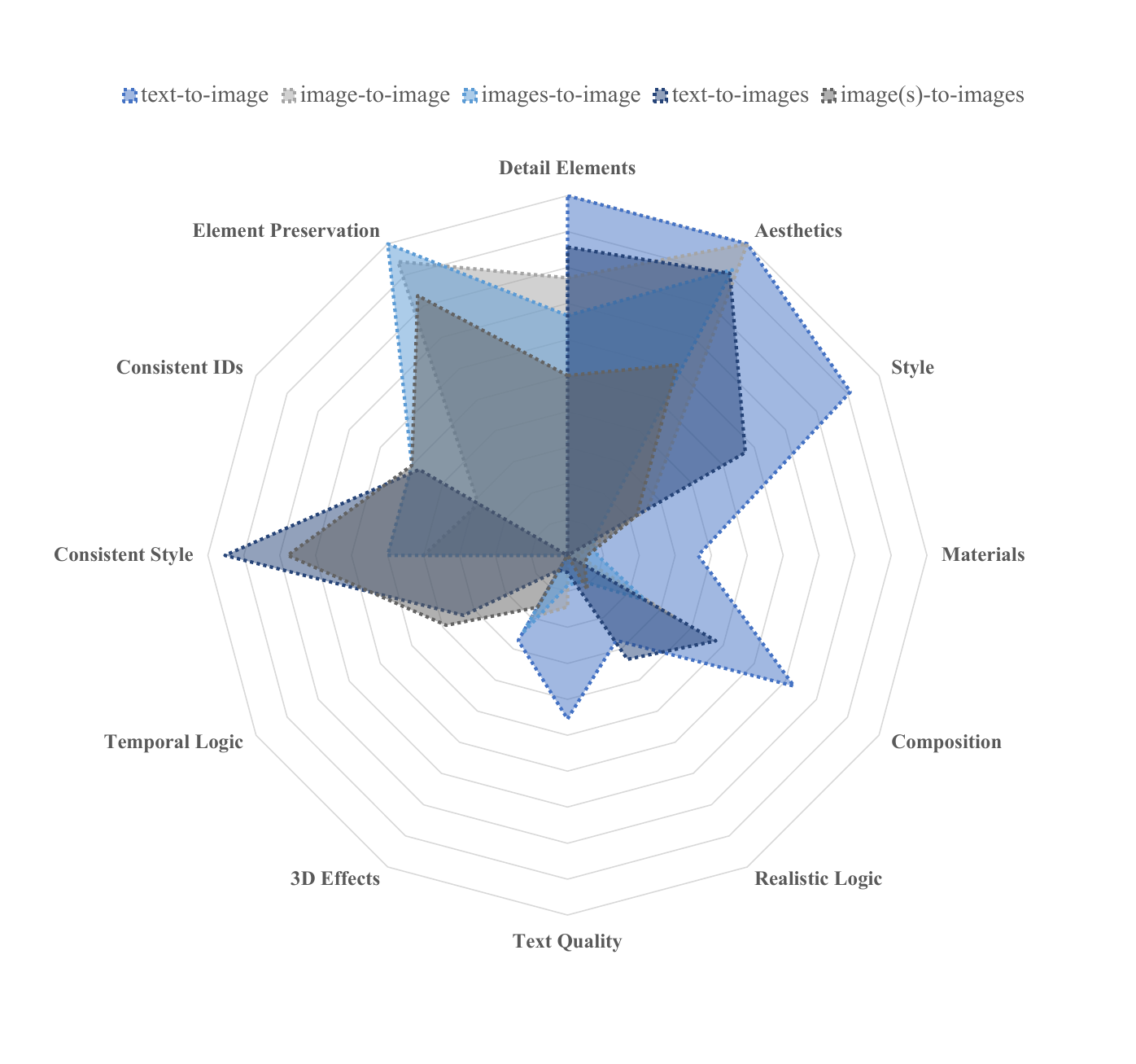}
\caption{\textbf{Statistics of evaluation dimensions for all tasks in \BenchName.} Each of the five task categories is represented by a distinct color. A total of 12 evaluation dimensions are analyzed, with the radar chart values indicating the proportion of evaluation questions related to each dimension within each category.} 
\label{fig:eval-dimension}
\end{center}
\end{figure}

We conduct a statistical analysis of the evaluation dimensions involved in each subtask within every category, with the results illustrated in the figure. In \cref{fig:eval-dimension}, a higher value for a dimension indicates that the category places greater emphasis on assessing the model’s capabilities in that dimension. The analysis reveals that all five categories prioritize the evaluation of aesthetic aspects and the quality of the association between the generated images and the details in the prompts. Specifically, \textbf{text-to-image} tasks emphasize assessments of style, image composition, and text quality. In contrast, \textbf{image-to-image} and \textbf{images-to-image} tasks focus on evaluating the retention of elements between the input and output images. Meanwhile, \textbf{text-to-images} and \textbf{image(s)-to-images} tasks, which involve generating multiple images, concentrate on evaluating dimensions such as ID consistency and style consistency among the generated images.

\section{Additional Experiments}

\paragraph{Supplementary results on image(s)-to-images}

Due to space constraints, we do not include all experimental results for the \textbf{image(s)-to-images} category in \cref{tab:is2is-score}. Supplementary results are provided in \cref{tab:is2is-cont}. The current abilities of all models to achieve inter-image associations like ID consistency and style consistency stem from GPT-4o’s \citep{gpt4o} detailed rephrasing of each prompt, akin to the group image descriptions in GDT \citep{huang2024group}. However, GDT employs a design where image tokens are concatenated during attention computation, whereas solely using MLLM rephrasing does not facilitate inter-image association modeling in the latent space. In the future, to enable multi-image generation tasks with complex associations, models will need to consider parallel generation of multiple images or utilize partially generated images as input conditions to guide the generation of subsequent images.

\begin{table*}[!t]
\caption{\textbf{Experimental results on Image(s)-to-Images.} Each task category is averaged across all its subtasks, with the top-ranked model scores for each task type highlighted in \textbf{bold}. Task types that a model cannot support are marked with "--". "$\dagger$" represents the use of MLLM for prompt rephrasing.}
\resizebox{\textwidth}{!}{
\begin{tabular}{lccccccccc}
\toprule
\multirow{2}{*}{\textbf{Method}} &
  \multicolumn{7}{c}{\textbf{Subtasks Score}} &
  \multirow{2}{*}{\textbf{Avg. Score}} \\ \cline{2-8}
   &
  Paint. Undo &
  Same Pose &
  Three-view Trans. &
  Child. Book &
  Plant Growth &
  Prod. Usage Scen. &
  Stop-motion Anim. &
   \\ \midrule \midrule
\rowcolor{gray!20}
FLUX-1$\dagger$ \citep{flux1} & 0.00 & 0.00 & 0.00 & \textbf{45.83} & 41.67 & \textbf{33.33} & \textbf{25.00} & \textbf{29.17} \\
DALL-E 3$\dagger$ \citep{ramesh2022hierarchical} & 0.00 & 0.00 & 0.00 & 37.50 & \textbf{58.30} & 16.67 & 16.67 & 14.44 \\
Stable Diffusion 3$\dagger$ \citep{esser2024scalingrectifiedflowtransformers} & 0.00 & \textbf{25.00} & 0.00 & 29.17 & 16.67 & 16.67 & 16.67 & 13.06 \\
Pixart$\dagger$ \citep{chen2023pixart} & 0.00 & 8.33 & 0.00 & 37.50 & 41.67 & 16.67 & 16.67 & 21.39 \\
InstructPix2Pix \citep{brooks2023instructpix2pix} & -- & -- & -- & -- & -- & -- & -- & -- \\
MagicBrush \citep{zhang2024magicbrush} & -- & -- & -- & -- & -- & -- & -- & -- \\
Anole \citep{chern2024anole} & 0.00 & 0.00 & 0.00 & 0.00 & 0.00 & 0.00 & 0.00 & 0.00 \\
Emu2 \citep{sun2024generative} & -- & -- & -- & -- & -- & -- & -- & -- \\
OmniGen \citep{xiao2024omnigen} & -- & -- & -- & -- & -- & -- & -- & -- \\ \bottomrule
\end{tabular}}
\label{tab:is2is-cont}
\end{table*}

\paragraph{GPT or Gemini for Evaluation?}

On a subset of the dataset, we select Gemini 1.5 Pro \citep{team2023gemini} to score the images generated by models based on the refined evaluation questions. However, MLLMs produce free-form textual outputs, making it challenging to ensure binary scores of 0 or 1 as human annotators do, potentially resulting in scoring failures. In \cref{tab:gpt-eval}, we report the failure rates of Gemini 1.5 pro \citep{team2023gemini} and GPT-4o \citep{gpt4o}, representing the proportion of evaluation questions where scoring failed. Specifically, we conduct three evaluations per question. If all three attempts do not yield a clear score, the evaluation is considered a failure. Across all models and evaluation questions, Gemini 1.5 pro exhibits a low failure rate of 0.95\%, whereas GPT-4o shows a high failure rate of 52.84\%, rendering it unsuitable as a reliable automated evaluation model. In practice, GPT-4o frequently responds with phrases such as \textit{"I'm sorry, I can’t assist with that"}, whereas Gemini 1.5 pro provides more consistent responses.



\begin{table*}[!t]
\centering
\caption{\textbf{Comparison of evaluation failure rates among different MLLMs.} For each evaluation question, MLLMs score the model-generated outputs three times. If none of the three scores return the required value (0 or 1), the evaluation is considered a failure.}
\resizebox{0.9\textwidth}{!}{
\begin{tabular}{lccccccccc}
\toprule
\multicolumn{1}{c}{\multirow{2}{*}{\textbf{Eval. MLLM}}} & \multicolumn{8}{c}{\textbf{Method}} & \multirow{2}{*}{\textbf{Total}} \\ \cline{2-9}
\multicolumn{1}{c}{} &
  \multicolumn{1}{l}{FLUX-1} &
  \multicolumn{1}{l}{DALL-E 3} &
  \multicolumn{1}{l}{SD3} &
  \multicolumn{1}{l}{Pixart} &
  \multicolumn{1}{l}{InstructPix2Pix} &
  \multicolumn{1}{l}{MagicBrush} &
  \multicolumn{1}{l}{Emu2} &
  \multicolumn{1}{l}{OmniGen} &
   \\ \midrule \midrule
\rowcolor{gray!20}
Gemini 1.5 pro~\citep{team2023gemini} & 0.33\% & 1.63\% & 1.96\% & 0.00\% & 0.00\% & 0.00\% & 0.67\% & 1.33\% & 0.95\% \\
GPT-4o~\citep{gpt4o}                  & 52.29\% & 54.58\% & 52.95\% & 56.86\% & 16.67\% & 26.67\% & 54.00\% & 53.33\% & 52.84\% \\ \bottomrule
\end{tabular}}
\label{tab:gpt-eval}
\end{table*}

\begin{table}[t]
\centering
\caption{\textbf{Additional experimental results on all categories of \BenchName.} "$\dagger$" represents the use of MLLM for prompt rephrasing. All models perform text-to-image generation on prompts rephrased by GPT-4o.}
\resizebox{0.7\columnwidth}{!}{
\begin{tabular}{lcccccc}
\toprule
\multirow{2}{*}{\textbf{Method}} & \multicolumn{5}{c}{\textbf{Scores on All Categories}} & \multirow{2}{*}{\textbf{Avg. Score}} \\ \cline{2-6}
                                          & T2I & I2I & Is2I & T2Is & I(s)2Is \\ \midrule \midrule
\rowcolor{gray!20}
FLUX-1$\dagger$ \citep{flux1}             & \textbf{46.06} & \textbf{12.13} & 4.89 & 20.15 & \textbf{29.17} & \textbf{22.48} \\
DALL-E 3$\dagger$ \citep{ramesh2022hierarchical}            & 24.34 & 6.95 & 5.26 & 14.36 & 14.44 & 13.07 \\
Stable Diffusion 3$\dagger$ \citep{esser2024scaling} & 24.04 & 10.79 & 4.69 & 21.59 & 13.06 & 14.83 \\
Pixart$\dagger$ \citep{chen2023pixart}             & 14.44 & 7.75 & 3.48 & 17.46 & 21.39 & 12.90 \\
Anole-T2I$\dagger$ \citep{chern2024anole}              & 0.00 & 3.10 & 1.26 & 8.98 & 8.89 & 4.45 \\
Emu2-T2I$\dagger$ \citep{sun2024generative}               & 17.98 & 3.15 & 2.34 & 15.53 & 12.78 & 10.36 \\
OmniGen-T2I$\dagger$ \citep{xiao2024omnigen}            & 21.41 & 6.09 & 4.50 & \textbf{23.52} & 21.39 & 15.38 \\ \bottomrule
\end{tabular}}
\label{tab:all-mllm}
\end{table}

\paragraph{Comparison of T2I capabilities across all models}

We also apply prompt rephrasing to all models in text-to-image generation settings. In this setup, all models have unified input comprehension capabilities, evaluating whether they can accurately translate prompts into high-quality generated images. The results are included in \cref{tab:all-mllm}, featuring Emu2 \citep{sun2024generative}, OmniGen \citep{xiao2024omnigen}, and Anole \citep{chern2024anole}, each distinguished by the “-T2I” suffix. FLUX-1 \citep{flux1} remains the top-ranked model. FLUX-1 demonstrates a strong ability to convert prompts into images, maintaining stable image quality with only rare instances of failure.


\section{Data Examples}

\begin{figure*}[!t]
\centering
\begin{center}
\includegraphics[width=1\linewidth]{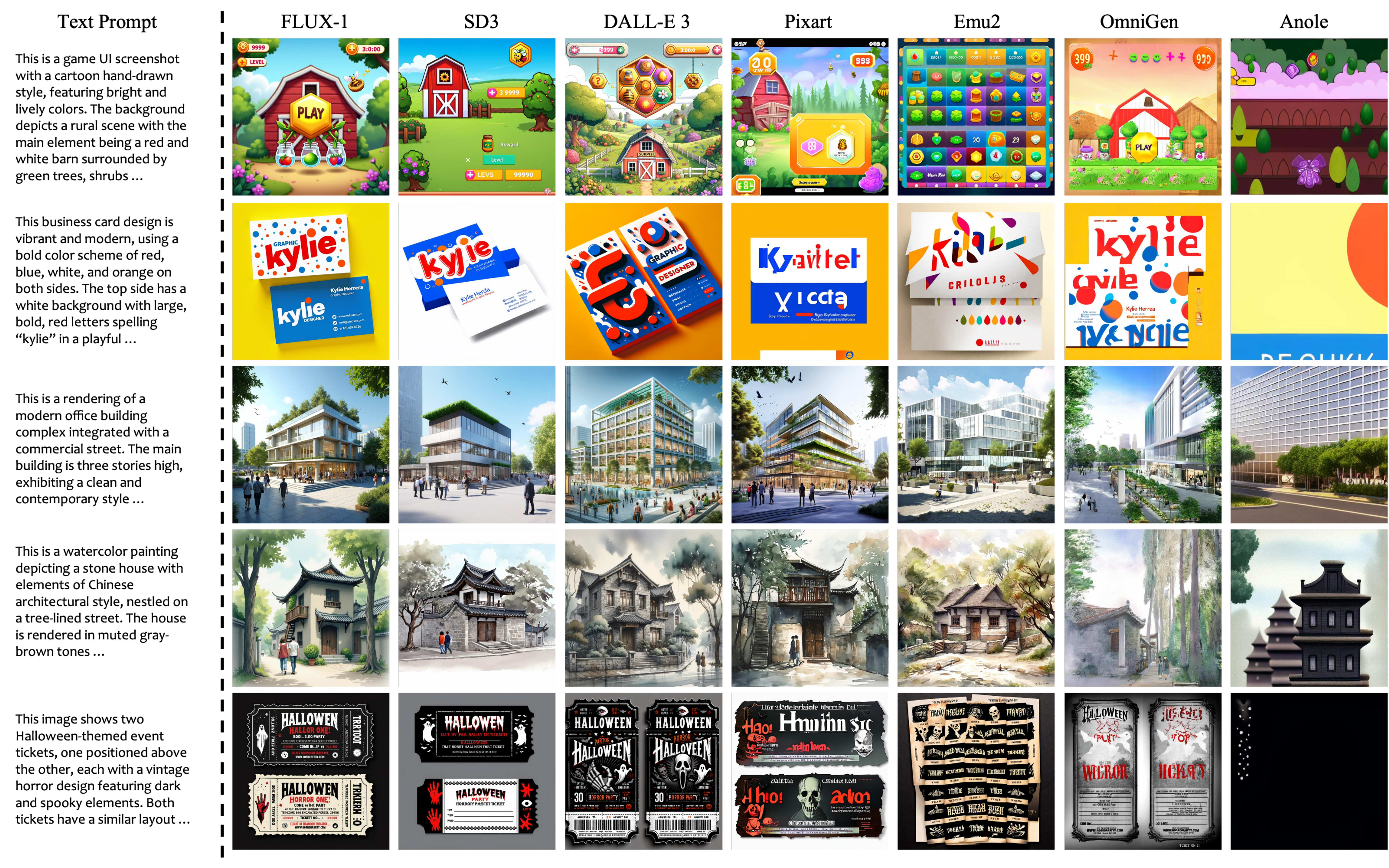}
\caption{\textbf{Generations for selected cases in the text-to-image category.} The displayed task categories, from top to bottom, include \textit{game UI generation}, \textit{business card generation}, \textit{architectural style generation}, \textit{painting generation}, and \textit{ticket generation}.} 
\label{fig:case-t2i}
\end{center}
\end{figure*}

\begin{figure*}[!t]
\centering
\begin{center}
\includegraphics[width=1\linewidth]{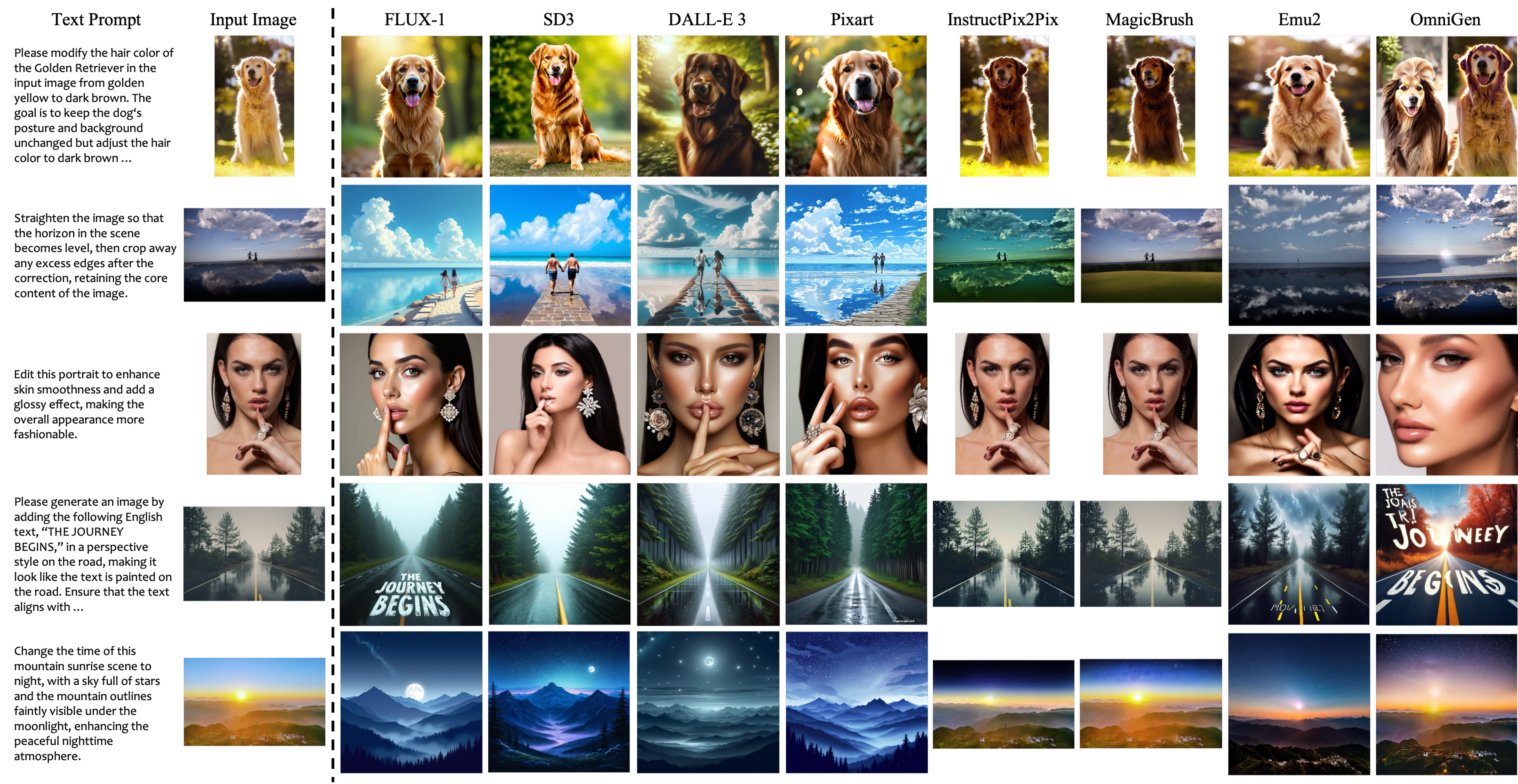}
\caption{\textbf{Generations for selected cases in the image-to-image category.} The displayed task categories, from top to bottom, include \textit{animal hair editing}, \textit{image straighten}, \textit{image retouching}, \textit{text insertion}, and \textit{time editing}.} 
\label{fig:case-i2i}
\end{center}
\end{figure*}

\begin{figure*}[!t]
\centering
\begin{center}
\includegraphics[width=1\linewidth]{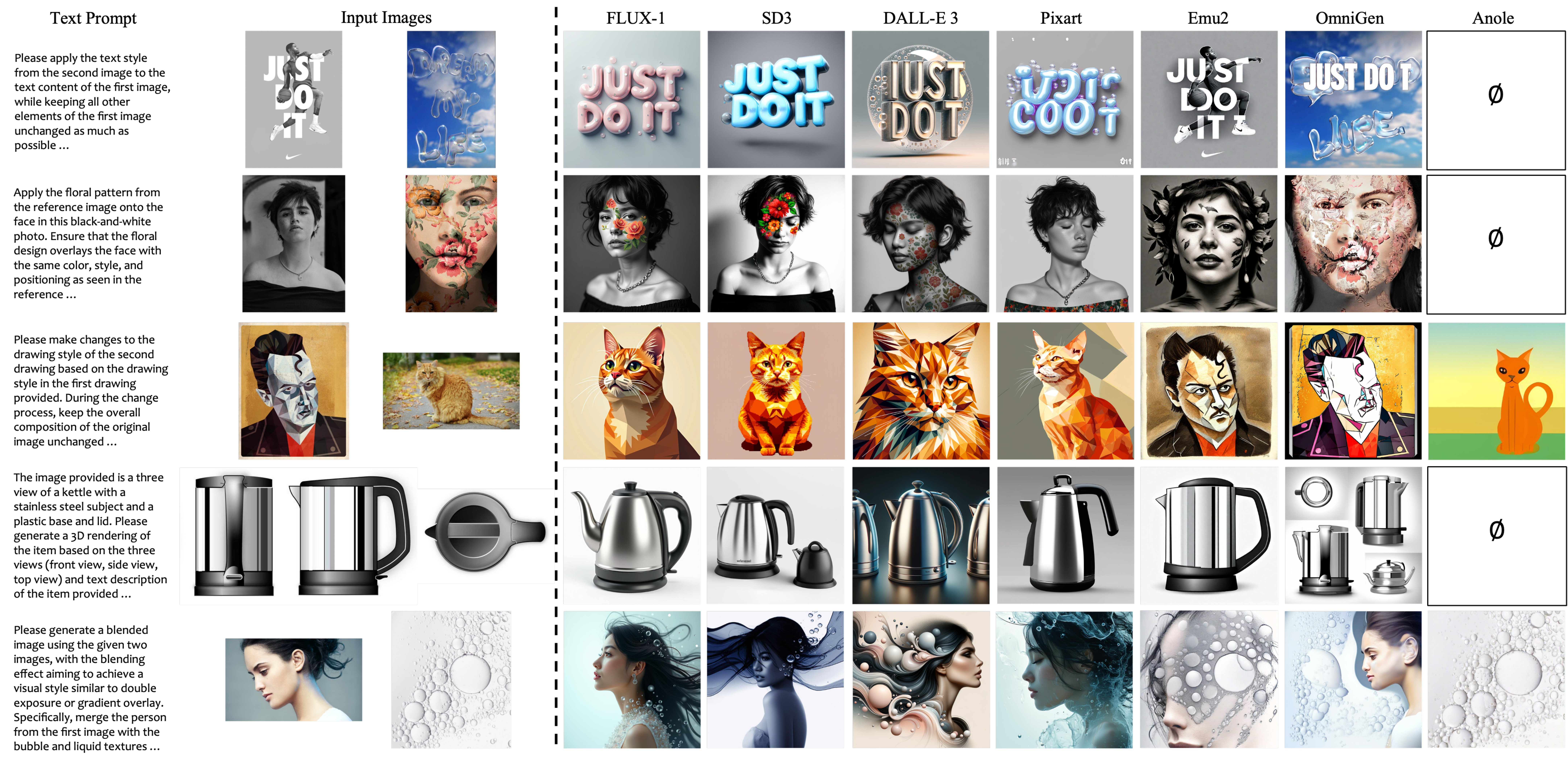}
\caption{\textbf{Generations for selected cases in the multi-image-to-image category.} The displayed task categories, from top to bottom, include \textit{text style transfer}, \textit{body painting transfer}, \textit{art style transfer}, \textit{3D rendering}, and \textit{double explosure}.} 
\label{fig:case-is2i}
\end{center}
\end{figure*}

\begin{figure*}[htbp]
    \centering
    \begin{subfigure}[b]{0.48\linewidth}
        \centering
        \includegraphics[width=\linewidth]{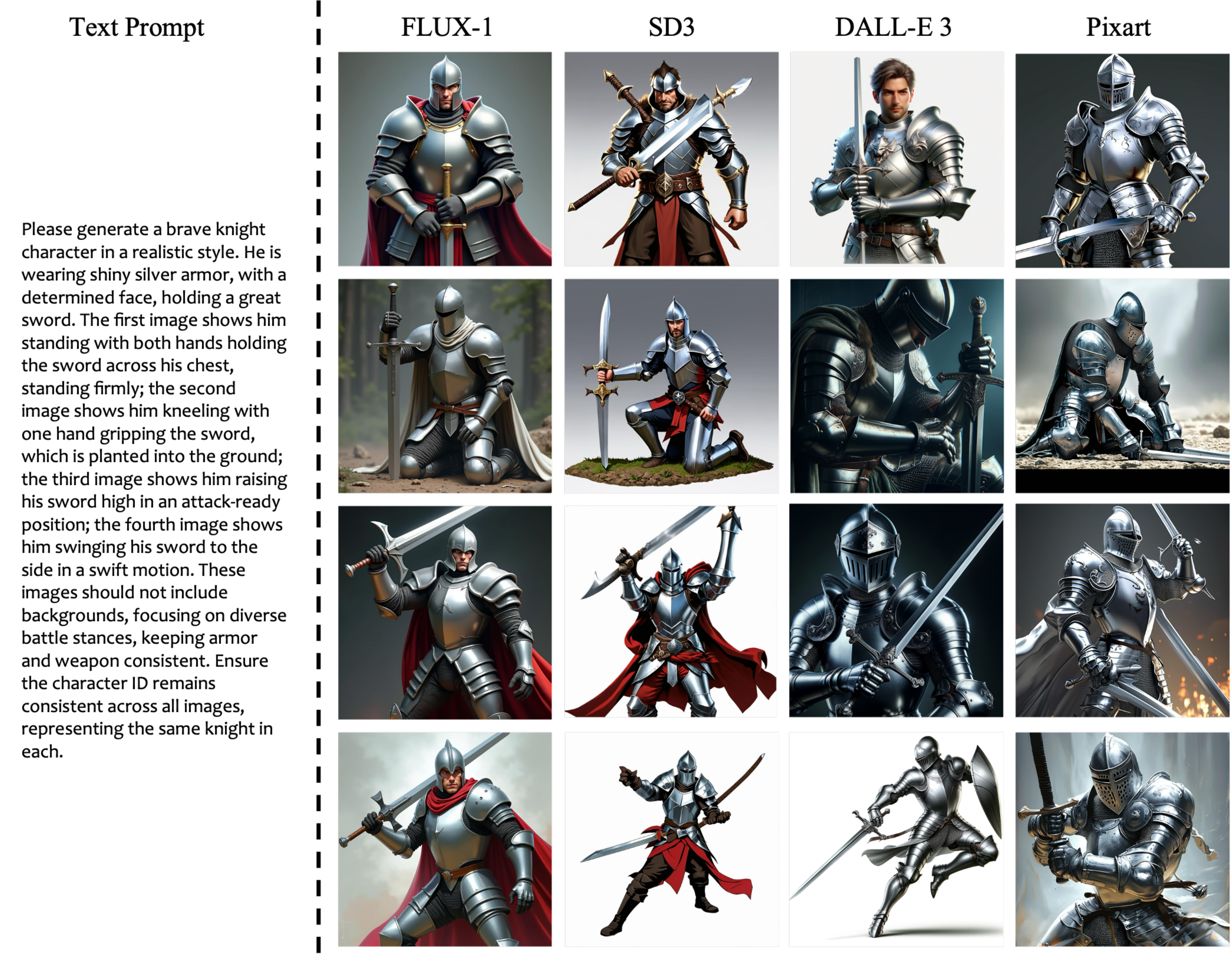}
        \caption{Generated images for \textit{character action design}}
        \label{fig:case-t2is-1}
    \end{subfigure}
    \hfill
    \begin{subfigure}[b]{0.48\linewidth}
        \centering
        \includegraphics[width=\linewidth]{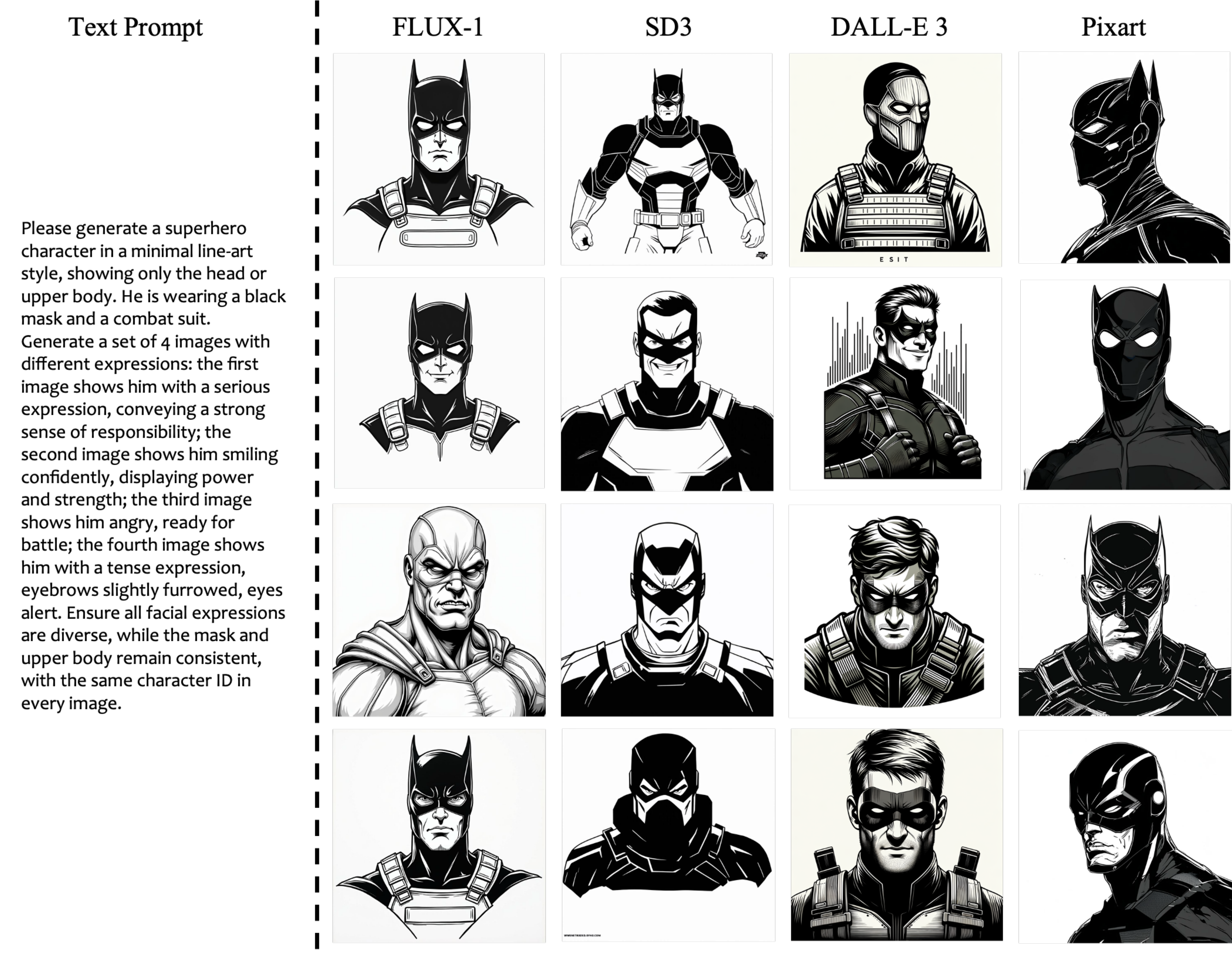}
        \caption{Generated images for \textit{character expression design}}
        \label{fig:case-t2is-2}
    \end{subfigure}
    \caption{\textbf{Generations for selected cases in the text-to-multi-image category.} }
    \label{fig:case-t2is}
\end{figure*}

\begin{figure*}[!t]
\centering
\begin{center}
\includegraphics[width=1\linewidth]{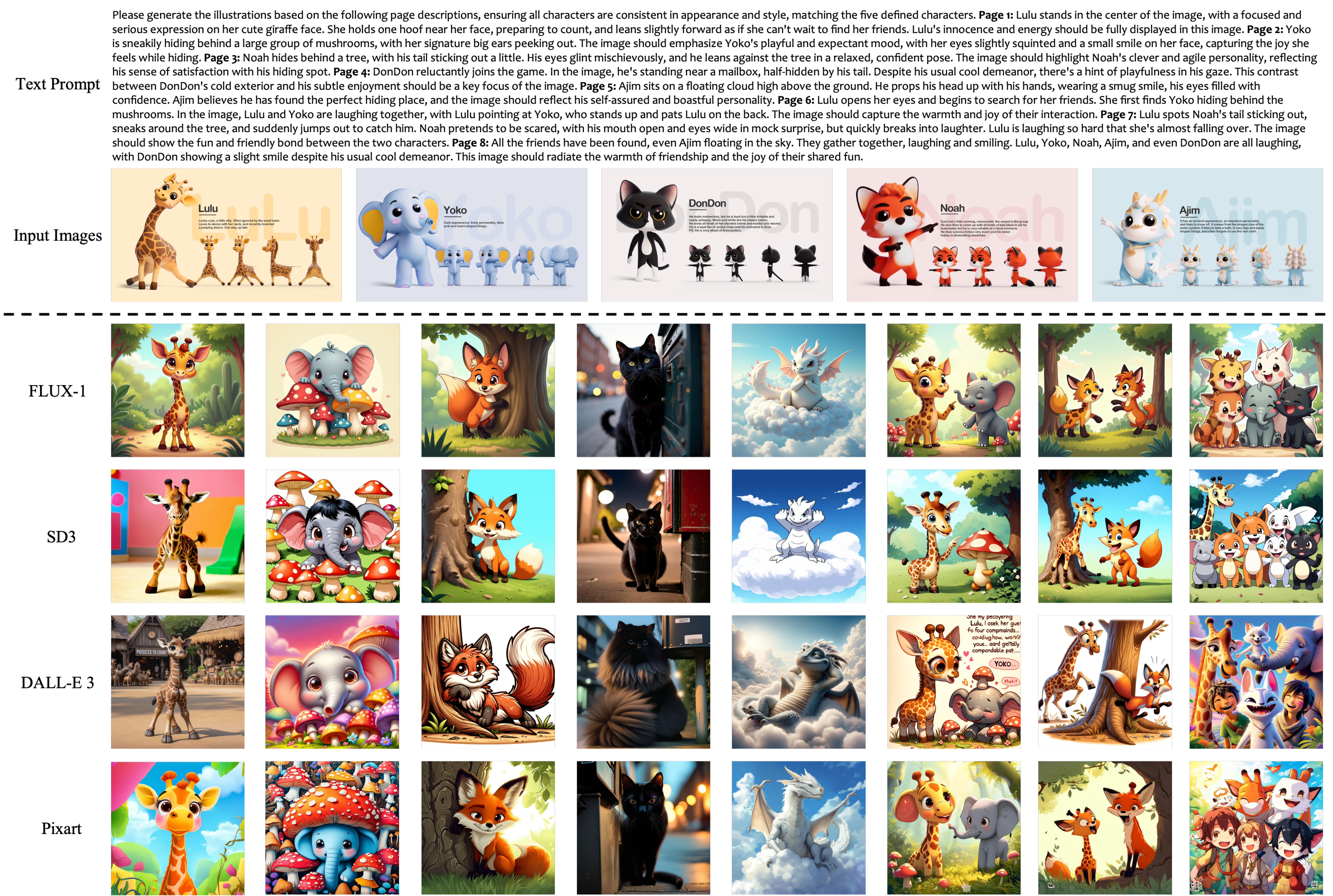}
\caption{\textbf{Generations for the task of \textit{children's storybook generation}.} The dashed line above represents the model’s input text prompts and role definition images, while the dashed line below illustrates the prompt generation results after rephrasing by GPT-4o \citep{gpt4o} for four models.} 
\label{fig:case-is2is}
\end{center}
\end{figure*}

Additional model-generated results are demonstrated in \cref{fig:case-t2i} - \cref{fig:case-is2is}, including the input images and text prompts used. Some text prompts are omitted due to their length. Since different models support a limited number of task categories, we only showcase the models that are capable of handling each respective category in the generation results.

\begin{figure*}[!t]
\centering
\begin{center}
\includegraphics[width=1\linewidth]{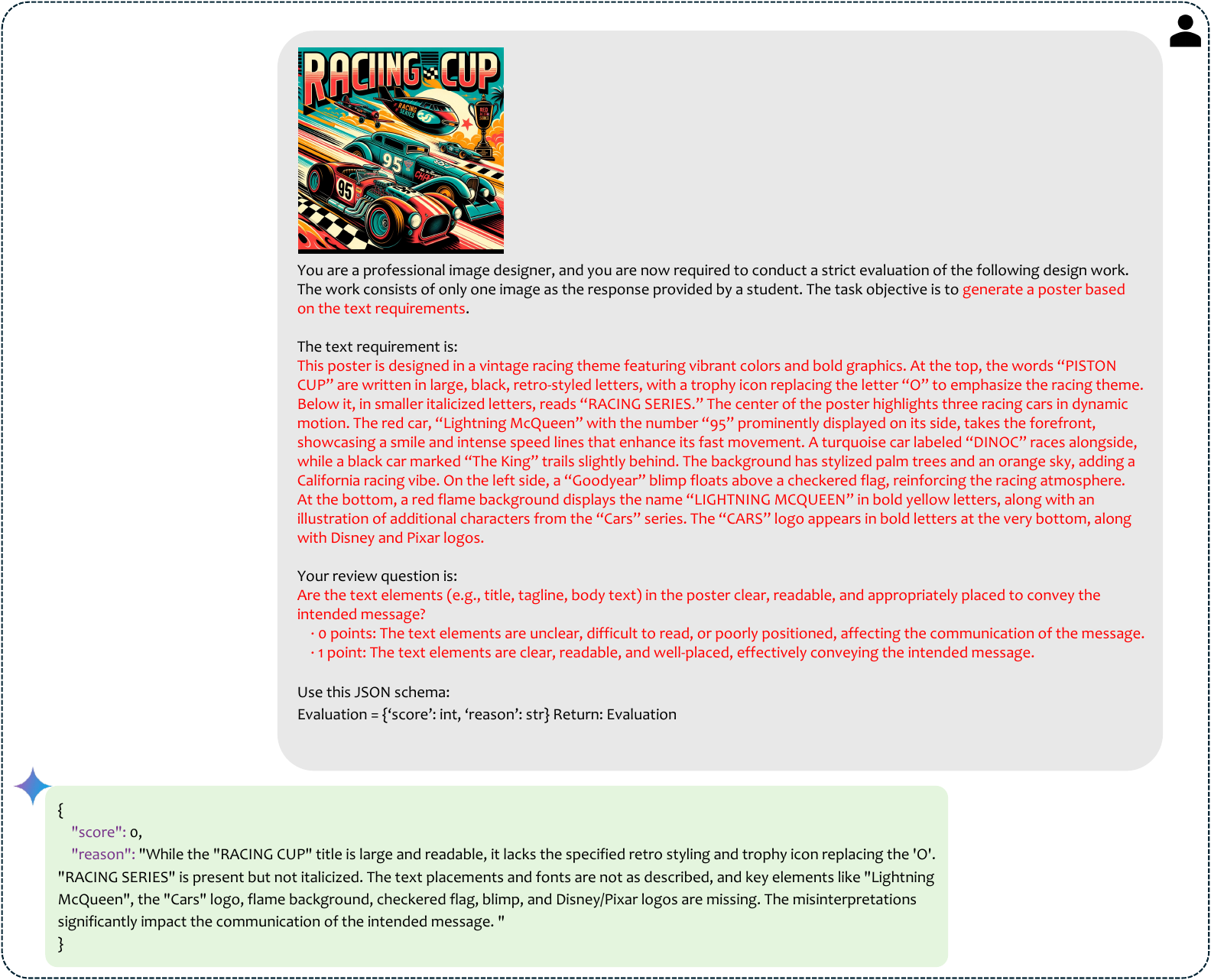}
\caption{\textbf{Automated evaluation of \textit{poster generation}.} The red font represents content that changes with each task or evaluation question.} 
\label{fig:eval-vis-1}
\end{center}
\end{figure*}

\begin{figure*}[!t]
\centering
\begin{center}
\includegraphics[width=1\linewidth]{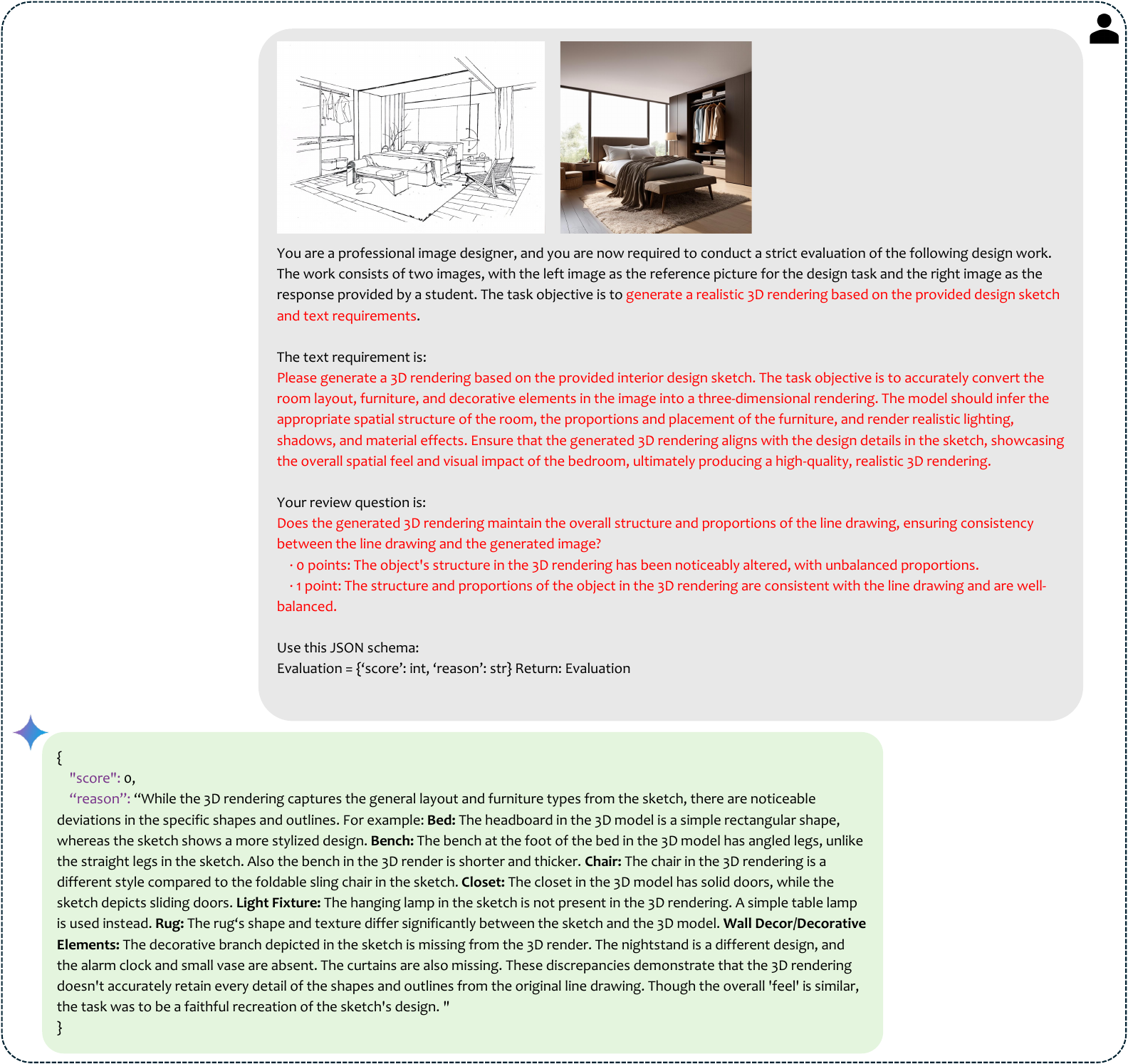}
\caption{\textbf{Automated evaluation of \textit{3D rendering generation}.} The red font represents content that changes with each task or evaluation question.} 
\label{fig:eval-vis-2}
\end{center}
\end{figure*}

Examples of automated evaluations conducted using Gemini 1.5 pro \citep{team2023gemini} are illustrated in \cref{fig:eval-vis-1} and \cref{fig:eval-vis-2}. Due to the detailed definitions of the generation prompts and evaluation questions, the evaluation process can be effectively transformed into a multimodal understanding task, which MLLM excels at. In both presented examples, the model-generated results fail to fully meet the prompt requirements, resulting in a score of 0.

\section{Limitations \& Future Work}

Due to the current capabilities of multimodal large language models (MLLMs) still falling short of human performance, we are unable to apply automated MLLM evaluations to all tasks while meeting the evaluation standards of professional designers. Furthermore, the primary goal of \BenchName~is to bridge the gap between current generative models and professional tasks, pushing model capabilities toward a professional level. However, there remains a significant distance to match the proficiency of professional designers. In the future, we will focus on updating and maintaining \BenchName, continuously refining automated evaluation methods in line with the real-time advancements of MLLMs, and expanding to more specialized tasks. This will ensure that the benchmark effectively supports the ongoing evolution of generative model capabilities.

%% file: main.bbl
\begin{thebibliography}{66}
\providecommand{\natexlab}[1]{#1}
\providecommand{\url}[1]{\texttt{#1}}
\expandafter\ifx\csname urlstyle\endcsname\relax
  \providecommand{\doi}[1]{doi: #1}\else
  \providecommand{\doi}{doi: \begingroup \urlstyle{rm}\Url}\fi

\bibitem[Ramesh et~al.(2021)Ramesh, Pavlov, Goh, Gray, Voss, Radford, Chen, and Sutskever]{ramesh2021zero}
Aditya Ramesh, Mikhail Pavlov, Gabriel Goh, Scott Gray, Chelsea Voss, Alec Radford, Mark Chen, and Ilya Sutskever.
\newblock Zero-shot text-to-image generation.
\newblock In \emph{International conference on machine learning}, pages 8821--8831. Pmlr, 2021.

\bibitem[Ramesh et~al.(2022)Ramesh, Dhariwal, Nichol, Chu, and Chen]{ramesh2022hierarchical}
Aditya Ramesh, Prafulla Dhariwal, Alex Nichol, Casey Chu, and Mark Chen.
\newblock Hierarchical text-conditional image generation with clip latents.
\newblock \emph{arXiv preprint arXiv:2204.06125}, 1\penalty0 (2):\penalty0 3, 2022.

\bibitem[Esser et~al.(2021)Esser, Rombach, and Ommer]{esser2021taming}
Patrick Esser, Robin Rombach, and Bjorn Ommer.
\newblock Taming transformers for high-resolution image synthesis.
\newblock In \emph{Proceedings of the IEEE/CVF conference on computer vision and pattern recognition}, pages 12873--12883, 2021.

\bibitem[Rombach et~al.(2022)Rombach, Blattmann, Lorenz, Esser, and Ommer]{rombach2022high}
Robin Rombach, Andreas Blattmann, Dominik Lorenz, Patrick Esser, and Bj{\"o}rn Ommer.
\newblock High-resolution image synthesis with latent diffusion models.
\newblock In \emph{Proceedings of the IEEE/CVF conference on computer vision and pattern recognition}, pages 10684--10695, 2022.

\bibitem[Saharia et~al.(2022)Saharia, Chan, Saxena, Li, Whang, Denton, Ghasemipour, Gontijo~Lopes, Karagol~Ayan, Salimans, et~al.]{saharia2022photorealistic}
Chitwan Saharia, William Chan, Saurabh Saxena, Lala Li, Jay Whang, Emily~L Denton, Kamyar Ghasemipour, Raphael Gontijo~Lopes, Burcu Karagol~Ayan, Tim Salimans, et~al.
\newblock Photorealistic text-to-image diffusion models with deep language understanding.
\newblock \emph{Advances in neural information processing systems}, 35:\penalty0 36479--36494, 2022.

\bibitem[Betker et~al.(2023)Betker, Goh, Jing, Brooks, Wang, Li, Ouyang, Zhuang, Lee, Guo, et~al.]{betker2023improving}
James Betker, Gabriel Goh, Li~Jing, Tim Brooks, Jianfeng Wang, Linjie Li, Long Ouyang, Juntang Zhuang, Joyce Lee, Yufei Guo, et~al.
\newblock Improving image generation with better captions.
\newblock \emph{Computer Science. https://cdn. openai. com/papers/dall-e-3. pdf}, 2\penalty0 (3):\penalty0 8, 2023.

\bibitem[Podell et~al.(2023)Podell, English, Lacey, Blattmann, Dockhorn, M{\"u}ller, Penna, and Rombach]{podell2023sdxl}
Dustin Podell, Zion English, Kyle Lacey, Andreas Blattmann, Tim Dockhorn, Jonas M{\"u}ller, Joe Penna, and Robin Rombach.
\newblock Sdxl: Improving latent diffusion models for high-resolution image synthesis.
\newblock \emph{arXiv preprint arXiv:2307.01952}, 2023.

\bibitem[Esser et~al.(2024{\natexlab{a}})Esser, Kulal, Blattmann, Entezari, M{\"u}ller, Saini, Levi, Lorenz, Sauer, Boesel, et~al.]{esser2024scaling}
Patrick Esser, Sumith Kulal, Andreas Blattmann, Rahim Entezari, Jonas M{\"u}ller, Harry Saini, Yam Levi, Dominik Lorenz, Axel Sauer, Frederic Boesel, et~al.
\newblock Scaling rectified flow transformers for high-resolution image synthesis.
\newblock In \emph{Forty-first International Conference on Machine Learning}, 2024{\natexlab{a}}.

\bibitem[Baldridge et~al.(2024)Baldridge, Bauer, Bhutani, Brichtova, Bunner, Chan, Chen, Dieleman, Du, Eaton-Rosen, et~al.]{baldridge2024imagen}
Jason Baldridge, Jakob Bauer, Mukul Bhutani, Nicole Brichtova, Andrew Bunner, Kelvin Chan, Yichang Chen, Sander Dieleman, Yuqing Du, Zach Eaton-Rosen, et~al.
\newblock Imagen 3.
\newblock \emph{arXiv preprint arXiv:2408.07009}, 2024.

\bibitem[Chen et~al.(2023)Chen, Yu, Ge, Yao, Xie, Wu, Wang, Kwok, Luo, Lu, et~al.]{chen2023pixart}
Junsong Chen, Jincheng Yu, Chongjian Ge, Lewei Yao, Enze Xie, Yue Wu, Zhongdao Wang, James Kwok, Ping Luo, Huchuan Lu, et~al.
\newblock Pixart-alpha: Fast training of diffusion transformer for photorealistic text-to-image synthesis.
\newblock \emph{arXiv preprint arXiv:2310.00426}, 2023.

\bibitem[Labs(2024{\natexlab{a}})]{blackforestlabs_flux_2024}
Black~Forest Labs.
\newblock Flux: Inference repository.
\newblock \url{https://github.com/black-forest-labs/flux}, 2024{\natexlab{a}}.
\newblock Accessed: 2024-10-25.

\bibitem[Zhang et~al.(2023{\natexlab{a}})Zhang, Rao, and Agrawala]{zhang2023adding}
Lvmin Zhang, Anyi Rao, and Maneesh Agrawala.
\newblock Adding conditional control to text-to-image diffusion models.
\newblock In \emph{Proceedings of the IEEE/CVF International Conference on Computer Vision}, pages 3836--3847, 2023{\natexlab{a}}.

\bibitem[Mou et~al.(2024)Mou, Wang, Xie, Wu, Zhang, Qi, and Shan]{mou2024t2i}
Chong Mou, Xintao Wang, Liangbin Xie, Yanze Wu, Jian Zhang, Zhongang Qi, and Ying Shan.
\newblock T2i-adapter: Learning adapters to dig out more controllable ability for text-to-image diffusion models.
\newblock In \emph{Proceedings of the AAAI Conference on Artificial Intelligence}, volume~38, pages 4296--4304, 2024.

\bibitem[Brooks et~al.(2023)Brooks, Holynski, and Efros]{brooks2023instructpix2pix}
Tim Brooks, Aleksander Holynski, and Alexei~A Efros.
\newblock Instructpix2pix: Learning to follow image editing instructions.
\newblock In \emph{Proceedings of the IEEE/CVF Conference on Computer Vision and Pattern Recognition}, pages 18392--18402, 2023.

\bibitem[Sheynin et~al.(2024)Sheynin, Polyak, Singer, Kirstain, Zohar, Ashual, Parikh, and Taigman]{sheynin2024emu}
Shelly Sheynin, Adam Polyak, Uriel Singer, Yuval Kirstain, Amit Zohar, Oron Ashual, Devi Parikh, and Yaniv Taigman.
\newblock Emu edit: Precise image editing via recognition and generation tasks.
\newblock In \emph{Proceedings of the IEEE/CVF Conference on Computer Vision and Pattern Recognition}, pages 8871--8879, 2024.

\bibitem[Radford et~al.(2019)Radford, Wu, Child, Luan, Amodei, Sutskever, et~al.]{radford2019language}
Alec Radford, Jeffrey Wu, Rewon Child, David Luan, Dario Amodei, Ilya Sutskever, et~al.
\newblock Language models are unsupervised multitask learners.
\newblock \emph{OpenAI blog}, 1\penalty0 (8):\penalty0 9, 2019.

\bibitem[Raffel et~al.(2020)Raffel, Shazeer, Roberts, Lee, Narang, Matena, Zhou, Li, and Liu]{raffel2020exploring}
Colin Raffel, Noam Shazeer, Adam Roberts, Katherine Lee, Sharan Narang, Michael Matena, Yanqi Zhou, Wei Li, and Peter~J Liu.
\newblock Exploring the limits of transfer learning with a unified text-to-text transformer.
\newblock \emph{Journal of machine learning research}, 21\penalty0 (140):\penalty0 1--67, 2020.

\bibitem[Brown(2020)]{brown2020language}
Tom~B Brown.
\newblock Language models are few-shot learners.
\newblock \emph{arXiv preprint arXiv:2005.14165}, 2020.

\bibitem[Ouyang et~al.(2022)Ouyang, Wu, Jiang, Almeida, Wainwright, Mishkin, Zhang, Agarwal, Slama, Ray, et~al.]{ouyang2022training}
Long Ouyang, Jeffrey Wu, Xu~Jiang, Diogo Almeida, Carroll Wainwright, Pamela Mishkin, Chong Zhang, Sandhini Agarwal, Katarina Slama, Alex Ray, et~al.
\newblock Training language models to follow instructions with human feedback.
\newblock \emph{Advances in neural information processing systems}, 35:\penalty0 27730--27744, 2022.

\bibitem[Zhang et~al.(2022)Zhang, Roller, Goyal, Artetxe, Chen, Chen, Dewan, Diab, Li, Lin, et~al.]{zhang2022opt}
Susan Zhang, Stephen Roller, Naman Goyal, Mikel Artetxe, Moya Chen, Shuohui Chen, Christopher Dewan, Mona Diab, Xian Li, Xi~Victoria Lin, et~al.
\newblock Opt: Open pre-trained transformer language models.
\newblock \emph{arXiv preprint arXiv:2205.01068}, 2022.

\bibitem[Touvron et~al.(2023{\natexlab{a}})Touvron, Lavril, Izacard, Martinet, Lachaux, Lacroix, Rozi{\`e}re, Goyal, Hambro, Azhar, et~al.]{touvron2023llama}
Hugo Touvron, Thibaut Lavril, Gautier Izacard, Xavier Martinet, Marie-Anne Lachaux, Timoth{\'e}e Lacroix, Baptiste Rozi{\`e}re, Naman Goyal, Eric Hambro, Faisal Azhar, et~al.
\newblock Llama: Open and efficient foundation language models.
\newblock \emph{arXiv preprint arXiv:2302.13971}, 2023{\natexlab{a}}.

\bibitem[Touvron et~al.(2023{\natexlab{b}})Touvron, Martin, Stone, Albert, Almahairi, Babaei, Bashlykov, Batra, Bhargava, Bhosale, et~al.]{touvron2023llama2}
Hugo Touvron, Louis Martin, Kevin Stone, Peter Albert, Amjad Almahairi, Yasmine Babaei, Nikolay Bashlykov, Soumya Batra, Prajjwal Bhargava, Shruti Bhosale, et~al.
\newblock Llama 2: Open foundation and fine-tuned chat models.
\newblock \emph{arXiv preprint arXiv:2307.09288}, 2023{\natexlab{b}}.

\bibitem[Dubey et~al.(2024)Dubey, Jauhri, Pandey, Kadian, Al-Dahle, Letman, Mathur, Schelten, Yang, Fan, et~al.]{dubey2024llama}
Abhimanyu Dubey, Abhinav Jauhri, Abhinav Pandey, Abhishek Kadian, Ahmad Al-Dahle, Aiesha Letman, Akhil Mathur, Alan Schelten, Amy Yang, Angela Fan, et~al.
\newblock The llama 3 herd of models.
\newblock \emph{arXiv preprint arXiv:2407.21783}, 2024.

\bibitem[Li et~al.(2024)Li, Zhang, Zhang, Guo, Zhang, Li, Zhang, Liu, and Li]{li2024llavanext-strong}
Bo~Li, Kaichen Zhang, Hao Zhang, Dong Guo, Renrui Zhang, Feng Li, Yuanhan Zhang, Ziwei Liu, and Chunyuan Li.
\newblock Llava-next: Stronger llms supercharge multimodal capabilities in the wild, May 2024.
\newblock URL \url{https://llava-vl.github.io/blog/2024-05-10-llava-next-stronger-llms/}.

\bibitem[Wang et~al.(2024{\natexlab{a}})Wang, Bai, Tan, Wang, Fan, Bai, Chen, Liu, Wang, Ge, et~al.]{wang2024qwen2}
Peng Wang, Shuai Bai, Sinan Tan, Shijie Wang, Zhihao Fan, Jinze Bai, Keqin Chen, Xuejing Liu, Jialin Wang, Wenbin Ge, et~al.
\newblock Qwen2-vl: Enhancing vision-language model's perception of the world at any resolution.
\newblock \emph{arXiv preprint arXiv:2409.12191}, 2024{\natexlab{a}}.

\bibitem[OpenAI(2024)]{gpt4o}
OpenAI.
\newblock Gpt-4o, 2024.
\newblock URL \url{https://openai.com/index/hello-gpt-4o/}.

\bibitem[Team et~al.(2023)Team, Anil, Borgeaud, Alayrac, Yu, Soricut, Schalkwyk, Dai, Hauth, Millican, et~al.]{team2023gemini}
Gemini Team, Rohan Anil, Sebastian Borgeaud, Jean-Baptiste Alayrac, Jiahui Yu, Radu Soricut, Johan Schalkwyk, Andrew~M Dai, Anja Hauth, Katie Millican, et~al.
\newblock Gemini: a family of highly capable multimodal models.
\newblock \emph{arXiv preprint arXiv:2312.11805}, 2023.

\bibitem[Ge et~al.(2023)Ge, Zhao, Zeng, Ge, Li, Wang, and Shan]{ge2023making}
Yuying Ge, Sijie Zhao, Ziyun Zeng, Yixiao Ge, Chen Li, Xintao Wang, and Ying Shan.
\newblock Making llama see and draw with seed tokenizer.
\newblock \emph{arXiv preprint arXiv:2310.01218}, 2023.

\bibitem[Sun et~al.(2023)Sun, Yu, Cui, Zhang, Zhang, Wang, Gao, Liu, Huang, and Wang]{sun2023emu}
Quan Sun, Qiying Yu, Yufeng Cui, Fan Zhang, Xiaosong Zhang, Yueze Wang, Hongcheng Gao, Jingjing Liu, Tiejun Huang, and Xinlong Wang.
\newblock Emu: Generative pretraining in multimodality.
\newblock In \emph{The Twelfth International Conference on Learning Representations}, 2023.

\bibitem[Sun et~al.(2024)Sun, Cui, Zhang, Zhang, Yu, Wang, Rao, Liu, Huang, and Wang]{sun2024generative}
Quan Sun, Yufeng Cui, Xiaosong Zhang, Fan Zhang, Qiying Yu, Yueze Wang, Yongming Rao, Jingjing Liu, Tiejun Huang, and Xinlong Wang.
\newblock Generative multimodal models are in-context learners.
\newblock In \emph{Proceedings of the IEEE/CVF Conference on Computer Vision and Pattern Recognition}, pages 14398--14409, 2024.

\bibitem[Wang et~al.(2024{\natexlab{b}})Wang, Zhang, Luo, Sun, Cui, Wang, Zhang, Wang, Li, Yu, et~al.]{wang2024emu3}
Xinlong Wang, Xiaosong Zhang, Zhengxiong Luo, Quan Sun, Yufeng Cui, Jinsheng Wang, Fan Zhang, Yueze Wang, Zhen Li, Qiying Yu, et~al.
\newblock Emu3: Next-token prediction is all you need.
\newblock \emph{arXiv preprint arXiv:2409.18869}, 2024{\natexlab{b}}.

\bibitem[Chern et~al.(2024)Chern, Su, Ma, and Liu]{chern2024anole}
Ethan Chern, Jiadi Su, Yan Ma, and Pengfei Liu.
\newblock Anole: An open, autoregressive, native large multimodal models for interleaved image-text generation.
\newblock \emph{arXiv preprint arXiv:2407.06135}, 2024.

\bibitem[Huang et~al.(2024)Huang, Wang, Wu, Dou, Shi, Feng, Liang, Liu, and Zhou]{huang2024group}
Lianghua Huang, Wei Wang, Zhi-Fan Wu, Huanzhang Dou, Yupeng Shi, Yutong Feng, Chen Liang, Yu~Liu, and Jingren Zhou.
\newblock Group diffusion transformers are unsupervised multitask learners.
\newblock \emph{arXiv preprint arXiv:2410.15027}, 2024.

\bibitem[Cho et~al.(2023)Cho, Zala, and Bansal]{cho2023dall}
Jaemin Cho, Abhay Zala, and Mohit Bansal.
\newblock Dall-eval: Probing the reasoning skills and social biases of text-to-image generation models.
\newblock In \emph{Proceedings of the IEEE/CVF International Conference on Computer Vision}, pages 3043--3054, 2023.

\bibitem[Ghosh et~al.(2024)Ghosh, Hajishirzi, and Schmidt]{ghosh2024geneval}
Dhruba Ghosh, Hannaneh Hajishirzi, and Ludwig Schmidt.
\newblock Geneval: An object-focused framework for evaluating text-to-image alignment.
\newblock \emph{Advances in Neural Information Processing Systems}, 36, 2024.

\bibitem[Hu et~al.(2023)Hu, Liu, Kasai, Wang, Ostendorf, Krishna, and Smith]{hu2023tifa}
Yushi Hu, Benlin Liu, Jungo Kasai, Yizhong Wang, Mari Ostendorf, Ranjay Krishna, and Noah~A Smith.
\newblock Tifa: Accurate and interpretable text-to-image faithfulness evaluation with question answering.
\newblock In \emph{Proceedings of the IEEE/CVF International Conference on Computer Vision}, pages 20406--20417, 2023.

\bibitem[Ruiz et~al.(2023)Ruiz, Li, Jampani, Pritch, Rubinstein, and Aberman]{ruiz2023dreambooth}
Nataniel Ruiz, Yuanzhen Li, Varun Jampani, Yael Pritch, Michael Rubinstein, and Kfir Aberman.
\newblock Dreambooth: Fine tuning text-to-image diffusion models for subject-driven generation.
\newblock In \emph{Proceedings of the IEEE/CVF conference on computer vision and pattern recognition}, pages 22500--22510, 2023.

\bibitem[Ku et~al.(2023)Ku, Li, Zhang, Lu, Fu, Zhuang, and Chen]{ku2023imagenhub}
Max Ku, Tianle Li, Kai Zhang, Yujie Lu, Xingyu Fu, Wenwen Zhuang, and Wenhu Chen.
\newblock Imagenhub: Standardizing the evaluation of conditional image generation models.
\newblock \emph{arXiv preprint arXiv:2310.01596}, 2023.

\bibitem[Yang et~al.(2024{\natexlab{a}})Yang, Ge, Li, Chen, Ge, Shan, and Chen]{yang2024seed}
Shuai Yang, Yuying Ge, Yang Li, Yukang Chen, Yixiao Ge, Ying Shan, and Yingcong Chen.
\newblock Seed-story: Multimodal long story generation with large language model.
\newblock \emph{arXiv preprint arXiv:2407.08683}, 2024{\natexlab{a}}.

\bibitem[Wu et~al.(2024{\natexlab{a}})Wu, Fei, Qu, Ji, and Chua]{wu24next}
Shengqiong Wu, Hao Fei, Leigang Qu, Wei Ji, and Tat-Seng Chua.
\newblock Next-gpt: Any-to-any multimodal llm.
\newblock In \emph{Proceedings of the International Conference on Machine Learning}, pages 53366--53397, 2024{\natexlab{a}}.

\bibitem[Chong and Forsyth(2020)]{chong2020effectively}
Min~Jin Chong and David Forsyth.
\newblock Effectively unbiased fid and inception score and where to find them.
\newblock In \emph{Proceedings of the IEEE/CVF conference on computer vision and pattern recognition}, pages 6070--6079, 2020.

\bibitem[Hessel et~al.(2021)Hessel, Holtzman, Forbes, Bras, and Choi]{hessel2021clipscore}
Jack Hessel, Ari Holtzman, Maxwell Forbes, Ronan~Le Bras, and Yejin Choi.
\newblock Clipscore: A reference-free evaluation metric for image captioning.
\newblock \emph{arXiv preprint arXiv:2104.08718}, 2021.

\bibitem[Esser et~al.(2024{\natexlab{b}})Esser, Kulal, Blattmann, Entezari, Müller, Saini, Levi, Lorenz, Sauer, Boesel, Podell, Dockhorn, English, Lacey, Goodwin, Marek, and Rombach]{esser2024scalingrectifiedflowtransformers}
Patrick Esser, Sumith Kulal, Andreas Blattmann, Rahim Entezari, Jonas Müller, Harry Saini, Yam Levi, Dominik Lorenz, Axel Sauer, Frederic Boesel, Dustin Podell, Tim Dockhorn, Zion English, Kyle Lacey, Alex Goodwin, Yannik Marek, and Robin Rombach.
\newblock Scaling rectified flow transformers for high-resolution image synthesis, 2024{\natexlab{b}}.
\newblock URL \url{https://arxiv.org/abs/2403.03206}.

\bibitem[Labs(2024{\natexlab{b}})]{flux1}
Black~Forest Labs.
\newblock Flux.1, 2024{\natexlab{b}}.
\newblock URL \url{https://github.com/black-forest-labs/flux}.

\bibitem[Zhang et~al.(2024{\natexlab{a}})Zhang, Mo, Chen, Sun, and Su]{zhang2024magicbrush}
Kai Zhang, Lingbo Mo, Wenhu Chen, Huan Sun, and Yu~Su.
\newblock Magicbrush: A manually annotated dataset for instruction-guided image editing.
\newblock \emph{Advances in Neural Information Processing Systems}, 36, 2024{\natexlab{a}}.

\bibitem[Gal et~al.(2022)Gal, Alaluf, Atzmon, Patashnik, Bermano, Chechik, and Cohen-Or]{gal2022image}
Rinon Gal, Yuval Alaluf, Yuval Atzmon, Or~Patashnik, Amit~H Bermano, Gal Chechik, and Daniel Cohen-Or.
\newblock An image is worth one word: Personalizing text-to-image generation using textual inversion.
\newblock \emph{arXiv preprint arXiv:2208.01618}, 2022.

\bibitem[Hu et~al.(2021)Hu, Shen, Wallis, Allen-Zhu, Li, Wang, Wang, and Chen]{hu2021lora}
Edward~J Hu, Yelong Shen, Phillip Wallis, Zeyuan Allen-Zhu, Yuanzhi Li, Shean Wang, Lu~Wang, and Weizhu Chen.
\newblock Lora: Low-rank adaptation of large language models.
\newblock \emph{arXiv preprint arXiv:2106.09685}, 2021.

\bibitem[Ruiz et~al.(2022)Ruiz, Li, Jampani, Pritch, Rubinstein, and Aberman]{ruiz2022dreambooth}
Nataniel Ruiz, Yuanzhen Li, Varun Jampani, Yael Pritch, Michael Rubinstein, and Kfir Aberman.
\newblock Dreambooth: Fine tuning text-to-image diffusion models for subject-driven generation.
\newblock \emph{arXiv}, 2022.

\bibitem[Team(2024)]{team2024chameleon}
Chameleon Team.
\newblock Chameleon: Mixed-modal early-fusion foundation models.
\newblock \emph{arXiv preprint arXiv:2405.09818}, 2024.

\bibitem[Zhou et~al.(2024)Zhou, Yu, Babu, Tirumala, Yasunaga, Shamis, Kahn, Ma, Zettlemoyer, and Levy]{zhou2024transfusionpredicttokendiffuse}
Chunting Zhou, Lili Yu, Arun Babu, Kushal Tirumala, Michihiro Yasunaga, Leonid Shamis, Jacob Kahn, Xuezhe Ma, Luke Zettlemoyer, and Omer Levy.
\newblock Transfusion: Predict the next token and diffuse images with one multi-modal model, 2024.

\bibitem[Xiao et~al.(2024)Xiao, Wang, Zhou, Yuan, Xing, Yan, Wang, Huang, and Liu]{xiao2024omnigen}
Shitao Xiao, Yueze Wang, Junjie Zhou, Huaying Yuan, Xingrun Xing, Ruiran Yan, Shuting Wang, Tiejun Huang, and Zheng Liu.
\newblock Omnigen: Unified image generation.
\newblock \emph{arXiv preprint arXiv:2409.11340}, 2024.

\bibitem[Petsiuk et~al.(2022)Petsiuk, Siemenn, Surbehera, Chin, Tyser, Hunter, Raghavan, Hicke, Plummer, Kerret, et~al.]{petsiuk2022human}
Vitali Petsiuk, Alexander~E Siemenn, Saisamrit Surbehera, Zad Chin, Keith Tyser, Gregory Hunter, Arvind Raghavan, Yann Hicke, Bryan~A Plummer, Ori Kerret, et~al.
\newblock Human evaluation of text-to-image models on a multi-task benchmark.
\newblock \emph{arXiv preprint arXiv:2211.12112}, 2022.

\bibitem[Bakr et~al.(2023)Bakr, Sun, Shen, Khan, Li, and Elhoseiny]{bakr2023hrs}
Eslam~Mohamed Bakr, Pengzhan Sun, Xiaoqian Shen, Faizan~Farooq Khan, Li~Erran Li, and Mohamed Elhoseiny.
\newblock Hrs-bench: Holistic, reliable and scalable benchmark for text-to-image models.
\newblock In \emph{Proceedings of the IEEE/CVF International Conference on Computer Vision}, pages 20041--20053, 2023.

\bibitem[Huang et~al.(2023)Huang, Sun, Xie, Li, and Liu]{huang2023t2i}
Kaiyi Huang, Kaiyue Sun, Enze Xie, Zhenguo Li, and Xihui Liu.
\newblock T2i-compbench: A comprehensive benchmark for open-world compositional text-to-image generation.
\newblock \emph{Advances in Neural Information Processing Systems}, 36:\penalty0 78723--78747, 2023.

\bibitem[Lee et~al.(2024)Lee, Yasunaga, Meng, Mai, Park, Gupta, Zhang, Narayanan, Teufel, Bellagente, et~al.]{lee2024holistic}
Tony Lee, Michihiro Yasunaga, Chenlin Meng, Yifan Mai, Joon~Sung Park, Agrim Gupta, Yunzhi Zhang, Deepak Narayanan, Hannah Teufel, Marco Bellagente, et~al.
\newblock Holistic evaluation of text-to-image models.
\newblock \emph{Advances in Neural Information Processing Systems}, 36, 2024.

\bibitem[Lin et~al.(2023)Lin, Yang, Li, Wang, and Wang]{lin2023designbench}
Kevin Lin, Zhengyuan Yang, Linjie Li, Jianfeng Wang, and Lijuan Wang.
\newblock Designbench: Exploring and benchmarking dall-e 3 for imagining visual design.
\newblock \emph{arXiv preprint arXiv:2310.15144}, 2023.

\bibitem[Zhao et~al.(2024)Zhao, Jin, Wang, and You]{zhao2024real}
Xuanlei Zhao, Xiaolong Jin, Kai Wang, and Yang You.
\newblock Real-time video generation with pyramid attention broadcast.
\newblock \emph{arXiv preprint arXiv:2408.12588}, 2024.

\bibitem[Yang et~al.(2024{\natexlab{b}})Yang, Teng, Zheng, Ding, Huang, Xu, Yang, Hong, Zhang, Feng, et~al.]{yang2024cogvideox}
Zhuoyi Yang, Jiayan Teng, Wendi Zheng, Ming Ding, Shiyu Huang, Jiazheng Xu, Yuanming Yang, Wenyi Hong, Xiaohan Zhang, Guanyu Feng, et~al.
\newblock Cogvideox: Text-to-video diffusion models with an expert transformer.
\newblock \emph{arXiv preprint arXiv:2408.06072}, 2024{\natexlab{b}}.

\bibitem[Bai et~al.(2023)Bai, Yang, Bai, Wang, Zhang, Lin, Wang, Zhou, and Zhou]{bai2023touchstone}
Shuai Bai, Shusheng Yang, Jinze Bai, Peng Wang, Xingxuan Zhang, Junyang Lin, Xinggang Wang, Chang Zhou, and Jingren Zhou.
\newblock Touchstone: Evaluating vision-language models by language models.
\newblock \emph{arXiv preprint arXiv:2308.16890}, 2023.

\bibitem[Liu et~al.(2025)Liu, Duan, Zhang, Li, Zhang, Zhao, Yuan, Wang, He, Liu, et~al.]{liu2025mmbench}
Yuan Liu, Haodong Duan, Yuanhan Zhang, Bo~Li, Songyang Zhang, Wangbo Zhao, Yike Yuan, Jiaqi Wang, Conghui He, Ziwei Liu, et~al.
\newblock Mmbench: Is your multi-modal model an all-around player?
\newblock In \emph{European Conference on Computer Vision}, pages 216--233. Springer, 2025.

\bibitem[Li et~al.(2023)Li, Wang, Wang, Ge, Ge, and Shan]{li2023seed}
Bohao Li, Rui Wang, Guangzhi Wang, Yuying Ge, Yixiao Ge, and Ying Shan.
\newblock Seed-bench: Benchmarking multimodal llms with generative comprehension.
\newblock \emph{arXiv preprint arXiv:2307.16125}, 2023.

\bibitem[Zhang et~al.(2024{\natexlab{b}})Zhang, Zhang, Tian, Fu, Zhang, Wu, Li, Wang, Wen, Zhang, et~al.]{zhang2024mme}
Yi-Fan Zhang, Huanyu Zhang, Haochen Tian, Chaoyou Fu, Shuangqing Zhang, Junfei Wu, Feng Li, Kun Wang, Qingsong Wen, Zhang Zhang, et~al.
\newblock Mme-realworld: Could your multimodal llm challenge high-resolution real-world scenarios that are difficult for humans?
\newblock \emph{arXiv preprint arXiv:2408.13257}, 2024{\natexlab{b}}.

\bibitem[Wang et~al.(2022)Wang, Kordi, Mishra, Liu, Smith, Khashabi, and Hajishirzi]{wang2022self}
Yizhong Wang, Yeganeh Kordi, Swaroop Mishra, Alisa Liu, Noah~A Smith, Daniel Khashabi, and Hannaneh Hajishirzi.
\newblock Self-instruct: Aligning language models with self-generated instructions.
\newblock \emph{arXiv preprint arXiv:2212.10560}, 2022.

\bibitem[Wu et~al.(2024{\natexlab{b}})Wu, Yang, Li, Zhang, Liu, Guibas, Lin, and Wetzstein]{wu2024gpt}
Tong Wu, Guandao Yang, Zhibing Li, Kai Zhang, Ziwei Liu, Leonidas Guibas, Dahua Lin, and Gordon Wetzstein.
\newblock Gpt-4v (ision) is a human-aligned evaluator for text-to-3d generation.
\newblock In \emph{Proceedings of the IEEE/CVF Conference on Computer Vision and Pattern Recognition}, pages 22227--22238, 2024{\natexlab{b}}.

\bibitem[Zhang et~al.(2023{\natexlab{b}})Zhang, Lu, Wang, Yan, Yan, Qin, Wang, Yan, Wang, and Petzold]{zhang2023gpt}
Xinlu Zhang, Yujie Lu, Weizhi Wang, An~Yan, Jun Yan, Lianke Qin, Heng Wang, Xifeng Yan, William~Yang Wang, and Linda~Ruth Petzold.
\newblock Gpt-4v (ision) as a generalist evaluator for vision-language tasks.
\newblock \emph{arXiv preprint arXiv:2311.01361}, 2023{\natexlab{b}}.

\bibitem[Zheng et~al.(2023)Zheng, Chiang, Sheng, Zhuang, Wu, Zhuang, Lin, Li, Li, Xing, et~al.]{zheng2023judging}
Lianmin Zheng, Wei-Lin Chiang, Ying Sheng, Siyuan Zhuang, Zhanghao Wu, Yonghao Zhuang, Zi~Lin, Zhuohan Li, Dacheng Li, Eric Xing, et~al.
\newblock Judging llm-as-a-judge with mt-bench and chatbot arena.
\newblock \emph{Advances in Neural Information Processing Systems}, 36:\penalty0 46595--46623, 2023.

\end{thebibliography}
